\documentclass[preprint,12pt]{elsarticle}

\usepackage{amssymb}
\usepackage{amsmath}
\usepackage{graphicx}
\usepackage{algorithm,}
\usepackage{algorithmic}
\usepackage{booktabs}
\usepackage{multirow}
\usepackage{adjustbox}
\usepackage{amsfonts}
\usepackage{import}
\usepackage{array}
\usepackage{url}
\usepackage{arydshln}
\usepackage{subfigure}

\newcommand{\bl}[1]{\underline{\bf{#1}}}

\journal{Neurocomputing}

\begin{document}

\begin{frontmatter}



\title{Mitigating Domain Mismatch in Face Recognition Using Style Matching}


\author{Chun-Hsien~Lin}
\author{Bing-Fei~Wu}



\begin{abstract}
Despite outstanding performance on public benchmarks, face recognition still suffers due to domain mismatch between training (source) and testing (target) data. Furthermore, these domains are not shared classes, which complicates domain adaptation. Since this is also a fine-grained classification problem which does not strictly follow the low-density separation principle, conventional domain adaptation approaches do not resolve these problems. In this paper, we formulate domain mismatch in face recognition as a style mismatch problem for which we propose two methods. First, we design a domain discriminator with human-level judgment to mine target-like images in the training data to mitigate the domain gap. Second, we extract style representations in low-level feature maps of the backbone model, and match the style distributions of the two domains to find a common style representation. Evaluations on verification and open-set and closed-set identification protocols show that both methods yield good improvements, and that performance is more robust if they are combined. Our approach is competitive with related work, and its effectiveness is verified in a practical application.
\end{abstract}




\end{frontmatter}



\section{Introduction}

Face recognition is an efficient tool for identity
authentication. When implemented using deep learning, its performance is nearly
perfect on public benchmarks, and it can be used in applications as varied as social
media, surveillance, police, and military. However, as shown persuasively in
Phillips~\cite{phillips2017cross}, in real-world deployments, powerful face 
recognition models usually exhibit poor performance,
regardless of the amount of training data. 
This is primarily due to domain mismatch,
that is, the mismatch between the training data (the source domain)
and testing data (target domain) distributions.
Such mismatch can be caused by differences in
illumination, race, gender, or age. One solution is to fine-tune the model
on the target domain with supervised training, but labeling is expensive, and
it is impossible to collect labeled data that covers all
possible scenarios. The problem is thus how best to adapt the model by 
leveraging unlabeled data in the target domain. 

Although many studies have been conducted on this problem, most 
assume that the domains are shared classes. The premise of face recognition, however, is
the opposite. In general, identities in a scenario do not constitute a training
dataset, which leads to an unique problem in face recognition. Therefore, of the few
studies on domain mismatch for face recognition~\cite{luo2018deep, 
wang2020deep, wang2019racial, arachchilage2020ssdl,
sohn2017unsupervised, hong2017sspp, arachchilage2020clusterface}, most 
adopt domain adaptation approaches. According to Yosinski 
et~al.~\cite{yosinski2014transferable}, for convolutional neural networks (CNNs), 
as lower layers are more transferable, the main idea for classical
domain adaptation is to fine-tune parameters by aligning the feature
distributions of the two domains in higher layers. 
They achieve modest improvements under the 
low-density separation assumption. In contrast to objects,
the contour variation among human faces is low, which means that face
recognition is a kind of fine-grained classification problem. Since feature
distributions of unseen identities do not necessarily satisfy the low-density separation
assumption, domain mismatch in face recognition
should be accounted for at the texture level.

One technique for domain adaptation is to adapt source images to appear as if
drawn from the target domain~\cite{bousmalis2017unsupervised, tran2019gotta,
li2020model} and to fine-tune the model with the transferred images to reduce the
domain gap. With the advance of generative adversarial networks
(GAN)~\cite{goodfellow2014generative}, it is possible to synthesize precise facial
images~\cite{karras2017progressive, karras2019style, tewari2020stylerig}. Thus it would 
seem that GANs could be used to reduce face recognition domain mismatch,
but using GANs to transfer source images could lead to two problems.
First, if non-face images are generated, learning would be harmed. Second,
the styles in both domains should be diverse, but as style is not easily
defined, it is difficult to determine which target style a source image should be
transferred to. Transferring each source image to the styles of all target
images would incur a complexity of $O(N^2)$. The GAN must be large enough to
generate high-quality images, and there must be a sufficient number of identities and images in
the training data. This would however preclude the
transfer of all the source images in a reasonable amount of time.

In this paper, we propose a feasible method for style-based (or, equivalently, 
texture-level) model adaptation. We follow Zhang et~al.~\cite{zhang2018unreasonable} in adopting a domain
discriminator with human-level judgment to evaluate the similarity of a
source image to the target domain. Also, per Chu et~al.~\cite{chu2016selective},
Wu et~al.~\cite{wu2018adaptive}, and Liu et~al.~\cite{liu2019reinforced}, 
we utilize evaluations to up-weight
source images that are visually similar to target images. This adaptation
mechanism is called perceptual scoring (PS). The style of an image can be
represented by feature map statistics from lower to higher CNN 
layers~\cite{ulyanov2017improved, dumoulin2016learned,
huang2017arbitrary, karras2019style}. We adopt Sinkhorn 
divergence~\cite{cuturi2013sinkhorn, cuturi2014fast, aude2016stochastic,
feydy2019interpolating, genevay2019sample, genevay2019entropy,
chizat2020faster} to align the style distributions of the source and target domains 
as if
the convolutional layers are learned to produce a common style space in
which source and target images are visually similar. This results in shallow-layer 
feature maps which are domain-invariant. We term this approach
\emph{style matching} (SM). 

To evaluate the performance of the proposed approaches, we conduct evaluations on 
IJB-A/B/C~\cite{klare2015pushing, whitelam2017iarpa, maze2018iarpa} since
they provide comprehensive face verification, closed-set
identification, and open-set identification protocols, facilitating evaluations 
under various scenarios. 
The effectiveness of our approach is competitive with that of related work
on domain adaptation for face recognition.
Moreover, to understand whether the proposed approaches are capable of solving real world problems,
we apply them to a surveillance application in the library of National Chiao Tung University (NCTU).
The testing results show that the proposed approaches can improves a baseline model by a large margin.

The contributions in this research can be summarized as follows:

\begin{itemize}
  \item We conduct a comprehensive study of domain adaptation for face recognition,
  showing that most domain adaptation techniques do not apply
  to face recognition. If adaptation algorithms for face recognition 
  are designed at the texture level, domain mismatch can be formulated as a style mismatch problem. 
  \item We propose perceptual scoring with human-level judgment to
  enhance the learning of visually target-like source images. Style
  matching adapts the model by learning a common style space in which the
  style distributions of both domains are confused.
  \item We conduct comprehensive testing protocols to cover all possible applications of
  face recognition to evaluate models objectively. Experiment results
  demonstrate the competitiveness of the proposed approaches.
  \item We apply the proposed approaches to a real scenario, 
  and show the significant improvements compared with the baseline.
\end{itemize}

The rest of this paper is structured as follows: 
In Section~\ref{related-work}, we briefly review related work on face recognition and domain adaptation. 
Then, in Section~\ref{proposed-approach}, we describe in detail the proposed approaches, 
and in Section~\ref{evaluations} we explain the evaluation protocols for different scenarios. 
In Section~\ref{experiments}, 
we explain the experimental settings, analyze the testing results, and compare these with prior work. 
We conclude in Section~\ref{conclusion}.


\section{Related Work}
\label{related-work}

\subsection{Face Recognition}

After deep CNN-based face recognition~\cite{taigman2014deepface}
successfully surpassed conventional methods by a large margin on a famous
benchmark~\cite{LFWTech}, this became the major backbone of face 
recognition~\cite{masi2018deep}. DeepID~\cite{ouyang2014deepid,
ouyang2015deepid, ouyang2016deepid} improves classification loss with
contrastive loss~\cite{hadsell2006dimensionality} to learn a discriminative
embedding space. Triplet loss~\cite{schroff2015facenet} separates
positive pairs from negative samples via a distance measure, and is free
from classification loss. Learning embeddings without a classifier is known as
metric learning~\cite{kaya2019deep}. Such a learning strategy relies heavily on good
data mining methods. Schroff et~al.~\cite{schroff2015facenet} adopt hard
sample mining to learn informative triplets. Center loss~\cite{wen2016discriminative} 
reduces intra-class variation by
regressing features to their cluster centers. To further separate
inter-class samples, many margin penalty losses have been proposed, including
SphereFace~\cite{liu2017sphereface}, CosFace~\cite{wang2018cosface}, and
ArcFace~\cite{deng2019arcface}. As presented in Deng et~al.~\cite{deng2019arcface},
classification loss with a suitable margin penalty significantly improves
performance. Thus determining a good margin is an active research focus. 
Liu et~al.~\cite{liu2019adaptiveface} learn margins by
maximizing them in total loss. Fair loss~\cite{liu2019fair} involves using a policy
network trained by reinforcement learning to control margins according to the
samples in the current batch. Huang et~al.~\cite{huang2020curricularface} change
the loss adaptively by evaluating the relation between intra- and
inter-class similarities. 

Despite the nearly perfect accuracy achieved by these methods, their performance depends 
on huge labeled datasets to achieve better generalization, 
and thus methods in this line of research cannot be used to account for domain mismatch.


\subsection{Domain Adaptation}

The aim of domain adaptation is to leverage unlabeled data to fine-tune a
trained model for the target domain; source data labels are assumed to be
available~\cite{wang2018deep}. The main challenge with domain adaptation is domain
mismatch, that is, dissimilar distributions for the two domains, which degrades 
model performance in the target domain. To eliminate this domain gap, some
focus on modifying the classifier's decision boundary~\cite{chu2016selective}, 
while others focus on learning an embedding space in which the
distributions of the two domains are aligned~\cite{long2015learning, sun2016deep,
long2018transferable, wu2018adaptive, ganin2015unsupervised,
tzeng2017adversarial, ma2019deep}. 
Target-domain performance can be refined by observing the low-density separation principle. 
Since aligning distributions does not fully satisfy the assumption that domains are shared classes,
studies have been conducted on developing better distribution matching
methods. Maximum mean discrepancy (MMD) has been widely adopted for domain 
adaptation~\cite{long2015learning, long2018transferable}. By mapping features to
a reproducing kernel Hilbert space (RKHS), we can estimate the domain discrepancy
through the squared distance between kernel mean embeddings. In
Long et~al.~\cite{long2015learning, long2018transferable}, MMD is used to match 
feature distributions on deeper CNN layers. 
Yan et~al.~\cite{yan2019weighted} weight MMD by class information to better match the feature distributions. 
Although MMD is simple and effective, it tends to focus on matching high-density regions~\cite{feydy2019interpolating}. 
The success of GANs~\cite{goodfellow2014generative} has revealed adversarial
learning as another option~\cite{ganin2015unsupervised,
tzeng2017adversarial, ma2019deep}: 
a domain discriminator is created to judge the domain of each sample.
For training, this results in learning an embedding space to
fool the domain discriminator, which results in the distributions being
confused. Although adversarial learning based approaches achieve improved
performance, they still involve the training of extra parameters
(for the discriminator). Also, in many approaches, the two models are trained
interactively, which complicates the setting of proper hyperparameters for
training.


\subsection{Domain Adaptation for Face Recognition}

Face recognition is a fine-grained classification problem for which the classes
(identities) in the source and target domains are not the same. Hence, 
conventional domain adaptation methods cannot be used
for face recognition. There is a paucity of research on domain adaptation for
face recognition.

\subsubsection{Distribution Matching}
Distributions can be matched by MMD~\cite{luo2018deep} or
adversarial learning~\cite{sohn2017unsupervised, hong2017sspp} to produce a
distribution-confused embedding space. 
However, these approaches are not sufficient for face recognition
since inter-class embeddings may not be separable in target domain.

\subsubsection{Clustering}
To make up for the shortcomings of distribution matching, 
clustering is used to assign pseudo labels in the target domain. 
In Wang and Deng~\cite{wang2020deep}, 
a target-domain classifier is trained using pseudo labels. 
Following the low-density separation principle, 
Wang et~al.~\cite{wang2019racial} use mutual information loss to categorize target-domain embeddings. 
Leveraging frame consistency in video face recognition, 
Arachchilage and Izquierdo~\cite{arachchilage2020ssdl} 
better estimate positive and negative pairs of target images. 
After clustering and mining, the model is fine-tuned using a modified version of triplet loss. 
Similar work is described in Arachchilage and Izquierdo~\cite{arachchilage2020clusterface}. 
As clustering-based methods rely on the target domain priors,
hyperparameters must be determined before clustering,
which is difficult in practice.

\subsubsection{Instance Normalization}
In work on style transfer~\cite{ulyanov2017improved, dumoulin2016learned,
huang2017arbitrary, karras2019style}, instance normalization is used
for style normalization. This could also be a way to 
learn a domain-invariant model. 
Li et~al.~\cite{li2016revisiting} remove bias by taking the mean and standard deviation 
from the target domain as batch normalization parameters.
Qing et~al.~\cite{qing2018improve} combine instance and batch normalization (IBN)
to improve cross-domain recognition. 
Nam et~al.~\cite{nam2018batch} propose batch-instance normalization (BIN), summing 
instance and batch normalization with an optimal ratio; 
they thus yield superior results by learning the ratio between two operators.
In Qian et~al.~\cite{qian2019deep}, 
the IBN layer is adopted as the domain adaptation layer, and the ratio between batch 
and instance normalization is determined by the domain statistics.
Drawing from these studies, Jin et~al.~\cite{jin2021style} propose style
normalization and restitution (SNR), extracting domain-invariant features
and preserving task-relevant information to further improve performance.
However, these approaches necessitate changes to the network architecture that incur
a far greater computational complexity. 



\section{Proposed Approach}
\label{proposed-approach}

\begin{figure*}[!ht]
\centering
\includegraphics[width=\textwidth]{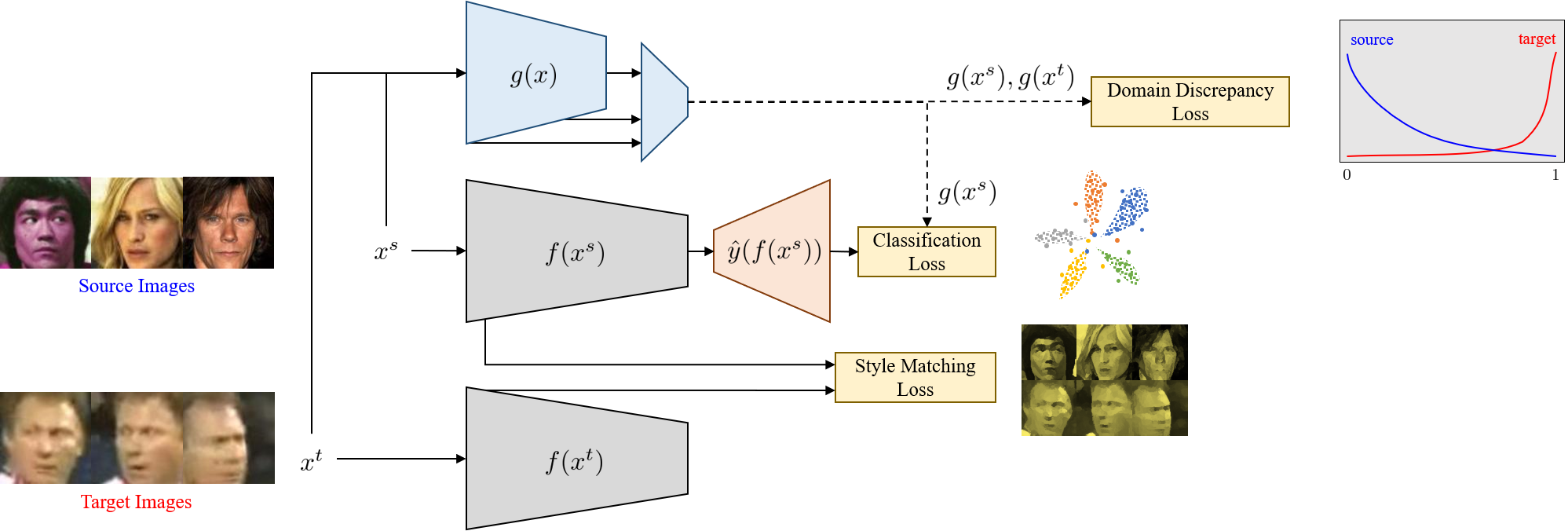}
\caption{Style-based domain adaptation. The domain
discriminator $g(x)$ is trained using domain discrepancy loss to score source images and
up-weight samples with higher scores in classification loss. Style features
are extracted from low-level feature maps. Style matching
loss finds a common style space in which the style distributions of two
domains are aligned.}
\label{fig:sheme}
\end{figure*}

With domain adaptation, the training dataset is taken as the source domain,
denoted by $X^s=\{x^s_i,y^s_i\}^{n^s}_{i=1}$, and the testing dataset is taken 
as the target domain, denoted by $X^t=\{x^t_j\}^{n^t}_{j=1}$. There are $n^s$ 
images in the source domain, and $n^t$ images in the target domain.
The source and target images are $x^s_i$ and $x^t_j$ respectively.
Each $x^s_i$ has a label $y^s_i$ which uses one-hot encoding.
The target image labels are unknown, reflecting the need for domain adaptation.
To simplify the notation, $x$, $x^s$, and $x^t$ stand for an
arbitrary image, a source image, and a target image respectively. 

In general face recognition, the source and target domain are not
shared classes, and the task is fine-grained categorization. In this case,
short feature distances between two images do not imply that they are visually similar.
We account for this by treating domain mismatch as style mismatch. 
We propose two approaches under different perspectives: perceptual scoring (PS) and style matching
(SM). We outline the proposed approach in Fig.~\ref{fig:sheme}. For PS we
draw from Zhang et~al.~\cite{zhang2018unreasonable} in the use of a domain discriminator, 
which judges the domain discrepancy of an image through a perceptual metric, 
to facilitate the learning of source images which are visually
similar to the target images. In SM, given a baseline face recognition model
$f(x)$, we extract style features from convolutional feature maps, and
align the distributions of style features from the two domains. Thus 
the face recognition model generates convolutional features with
homogeneous styles and focuses on learning target-like source images to improve
testing performance.

The rest of this section is structured as follows. In the next subsection,
we describe the primary classification loss with which we train the model by source
images with the supervision of labels. In the second and third
subsections, we describe the PS and SM methods respectively,
after which we summarize the two style-based domain adaptation methods.

\subsection{Classification Loss}

For visual recognition, the goal of an embedding function is to
encode images to an embedding space in which similar images are close together and
dissimilar images are far apart so that we can judge how similar two given
images are, or which category an image belongs to. The easiest way to train 
an embedding function is to use an additional classifier. The embedding
function is modeled by a convolutional neural network, $f(x)$, and the
classifier is a single-layer perceptron (a matrix) followed by softmax,
$\hat{y}(f(x))$. Since the output of the classifier is a probability
distribution, we train the cascaded model, $\hat{y}(f(x))$, by
minimizing cross entropy. The training loss is expressed as
\begin{equation}
\mathcal{L}_c = -y^s\log{\hat{y}(f(x^s))}
\label{eq:Lc}
\end{equation}
which is termed classification loss. We use this to train a baseline
model from scratch and preserve inter-class distances when adapting the model.


\subsection{Perceptual Scoring}

Assuming that domain information is shared between two domains,
we draw from Chu et~al.~\cite{chu2016selective}, Wu et~al.~\cite{wu2018adaptive}, 
and Liu et~al.~\cite{liu2019reinforced}
by mining informative images in the training dataset to mitigate domain shift. 

\subsubsection{Domain Discriminator}

\begin{figure}[!ht]
\centering
\includegraphics[width=3.0in]{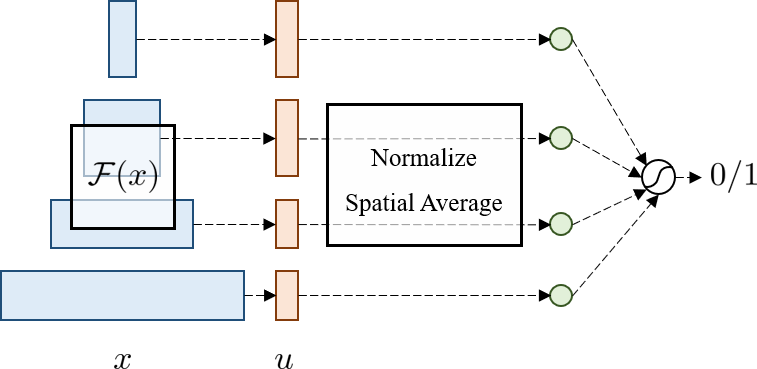}
\caption{Domain discriminator $g(x)$, modified from Zhang et~al.~\cite{zhang2018unreasonable}. 
The backbone $\mathcal{F}(x)$ of $g(x)$ is SqueezeNet~\cite{iandola2016squeezenet}.} 
\label{fig:lpips}
\end{figure}

The architecture from Zhang et~al.~\cite{zhang2018unreasonable} is modified as the
domain discriminator, as shown in Fig.~\ref{fig:lpips}. This discriminator 
approximates human-level judgments to calculate a score that represents
how similar an image is to the target domain.
For image~$x$, a feature
stack is extracted from $L_g$ layers of a backbone network $\mathcal{F}(x)$
and unit-normalized in the channel dimension, denoted by
$\mathcal{F}^l(x) \in \mathbb{R}^{H_l \times W_l \times C_l}$ for layer~$l$.
Each $\mathcal{F}^l(x)$ is scaled by vector $u^l \in \mathbb{R}^{C_l}$ and
summed channel-wise. We average spatially to yield a scalar representation $r^l
\in \mathbb{R}^1$ on each layer:
\begin{equation}
r^l = \frac{1}{H_lW_l}\sum_{h,w}{u^l \otimes \mathcal{F}^l_{hw}(x)}
\end{equation}
where $\otimes$ is a convolution operator. We then concatenate the
representations to produce a perceptual feature $r \in \mathbb{R}^{L_g}$, which we pass 
through a single-layer perceptron (vector~$v$) followed by a sigmoid activation
to judge the input image. The domain discriminator $g(x)$ is expressed as
\begin{equation}
g(x) = \frac{1}{1 + e^{-v^Tr}}.
\end{equation}
Note that since we use SqueezeNet~\cite{iandola2016squeezenet} as the backbone
network, in our case $L_g = 7$.

\subsubsection{Training and Adaptation}

We sample images from the source and target domains equally and use the
following loss function (domain discrepancy loss) to train the
domain discriminator:
\begin{equation}
\mathcal{L}_d = g^2(x^s) + (g(x^t) - 1)^2,
\end{equation}
in which we expect the discriminator to produce a low score (tending to 0) for a
source image but a high score (tending to 1) for a target image.
This score is used in the classification loss in
Equation~\ref{eq:Lc} to up-weight the loss caused by target-like images. The
classification loss is modified as
\begin{equation}
\mathcal{L}^{*}_c = g(x^s)\mathcal{L}_c,
\label{eq:wLc}
\end{equation}
where $\mathcal{L}^{*}_c$ is the loss for adapting the model based on PS.


\subsection{Style Matching}

Although PS indirectly reduces domain shift, its premise limits 
performance due to a lack of target-like images in the training dataset. To
compensate for this weakness, with SM we seek a common space in which the
style distributions of two domains are aligned.

\subsubsection{Style Measures}
According to studies on style transfer~\cite{ulyanov2017improved,
dumoulin2016learned, huang2017arbitrary, karras2019style}, the stacks of
mean and standard deviation on the feature maps describe the style of an
image. For layer~$l$ in the face recognition model, the style representation is
designated as $\{\mu(f^l(x)), \sigma(f^l(x))\}$. The measures of styles from
the two domains are defined as
\begin{equation}
p^l_{\mu} \overset{\textit{def.}}{=} \frac{1}{n^s}\sum_{i=1}^{n^s}\delta_{\mu(f^l(x^s_i))},~
p^l_{\sigma} \overset{\textit{def.}}{=} \frac{1}{n^s}\sum_{i=1}^{n^s}\delta_{\sigma(f^l(x^s_i))}
\end{equation}
\begin{equation}
q^l_{\mu} \overset{\textit{def.}}{=} \frac{1}{n^t}\sum_{j=1}^{n^t}\delta_{\mu(f^l(x^t_j))},~
q^l_{\sigma} \overset{\textit{def.}}{=} \frac{1}{n^t}\sum_{j=1}^{n^t}\delta_{\sigma(f^l(x^t_j))},
\end{equation}
where $p^l_\mu$ and $p^l_\sigma$ are measures of $\{\mu(f^l(x^s)),
\sigma(f^l(x^s))\}$, and $q^l_\mu$ and $q^l_\sigma$ are measures of
$\{\mu(f^l(x^t)), \sigma(f^l(x^t))\}$. We seek to minimize a certain
distribution matching loss between $\{p^l_\mu, p^l_\sigma\}$, and $\{q^l_\mu,
q^l_\sigma\}$.

\subsubsection{Matching Distribution}
In the literature on domain adaptation, MMD and adversarial learning are the main ways to match distributions. 
Although MMD is simple and necessitates no extra training parameters, it focuses on
matching the region of high density~\cite{feydy2019interpolating}, which may incur greater bias. 
Adversarial learning approaches, however, approximate relative entropy or
the Wasserstein distance (optimal transport) by training the embedding function and
the discriminator interactively. Although adversarial learning yields a
better estimation of domain discrepancy, it requires extra parametric subnetworks,
which complicates training. We strike a balance between the two methods by calculating
the distribution matching loss using Sinkhorn divergence~\cite{cuturi2013sinkhorn, cuturi2014fast,
aude2016stochastic, feydy2019interpolating, genevay2019sample,
genevay2019entropy, chizat2020faster}, which interpolates optimal transport and
MMD, and has been successfully applied to generative models in Genevay et~al.~\cite{genevay2018learning}. 

\subsubsection{Sinkhorn Divergence}
Similar to Genevay et~al.~\cite{genevay2018learning}, we align the style distributions
of the source and target domains by minimizing
\begin{equation}
\mathcal{W}_{\varepsilon}(\mathcal{P},\mathcal{Q}) = \underset{
    P \in \mathbb{R}^{n \times m}_{+};~
    P\mathbf{1}_m = \mathbf{1}_n;~
    P^T\mathbf{1}_n = \mathbf{1}_m
  }
  {\min} \langle{P, \hat{c}}\rangle,
\label{eq:r-wasserstein}
\end{equation}
where $\hat{c} \overset{\textit{def.}}{=} [ c(z^\mathcal{P}_i, z^\mathcal{Q}_j) ]_{i,j}
\in \mathbb{R}^{n \times m}$. Here, $z^\mathcal{P}_i$ and
$z^\mathcal{Q}_j$ are samples from distributions $\mathcal{P}$ and $\mathcal{Q}$
respectively. Where $p > 0$ is some exponent, 
$c(z^\mathcal{P}_i, z^\mathcal{Q}_j) = d(z^\mathcal{P}_i, z^\mathcal{Q}_j)^p$
is a cost equipped with a
distance function $d$. We define $d$ as Euclidean distance, and set $p=2$.
The entropy-regularized optimal transport, that is, Wasserstein distance,
is $\mathcal{W}_{\varepsilon}(\mathcal{P},\mathcal{Q})$, in which factor $\varepsilon$ 
regularizes entropy. Thus $P$ can be treated as a mass-moving plan between
$\mathcal{P}$ and $\mathcal{Q}$. As expressed in Cuturi~\cite{cuturi2013sinkhorn},
Cuturi and Doucet~\cite{cuturi2014fast}, and Genevay et~al.~\cite{genevay2018learning}, 
regularized optimal transport is
equivalent to restricting the search space by using the scaling form of $P$:
\begin{equation}
P_{i,j} = a_i K_{i,j} b_j,
\end{equation}
where $K_{i,j} \overset{\textit{def.}}{=} e^{\frac{-\hat{c}_{i,j}}{\varepsilon}}$;
the role of $\varepsilon$ here mirrors that of bandwidth in MMD. The optimization
problem of Equation~\ref{eq:r-wasserstein} turns into finding $a$ and $b$ that
meet certain requirements.
Based on the Sinkhorn algorithm~\cite{cuturi2013sinkhorn,
cuturi2014fast}, $a$ and $b$ can be found by
\begin{equation}
a_{l+1} \overset{\textit{def.}}{=} \frac{\mathbf{1}_n}{{K} b_{l}} ~ \quad\mathrm{and}\quad 
b_{l+1} \overset{\textit{def.}}{=} \frac{\mathbf{1}_m}{{K^T} a_{l+1}},
\end{equation}
where the fraction here is element-wise division. Starting with $b_0 =
\mathbf{1}_m$, we obtain feasible $a_L$ and $b_L$ after $L$ iterations. The
budget~$L$ is set to $10$ according to Genevay et~al.~\cite{genevay2018learning}. Hence,
Equation~\ref{eq:r-wasserstein} can be modified as
\begin{equation}
\mathcal{W}_{\varepsilon}(\mathcal{P},\mathcal{Q}) = \frac{1}{nm}a^T_L(K \odot \hat{c})b_L,
\end{equation}
where $\odot$ is an element-wise multiplication operator. Since $\hat{c}$ can
be large, to prevent numerical explosions during training, we divide by
$nm$ to normalize the total cost. However, such a discrepancy approximation
tends to overfit. To account for this, we add intra-measure
Wasserstein distance to penalize the inter-measure Wasserstein distance:
\begin{equation}
\tilde{\mathcal{W}}_{\varepsilon}(\mathcal{P},\mathcal{Q}) = 
  2\mathcal{W}_{\varepsilon}(\mathcal{P},\mathcal{Q})
  - \mathcal{W}_{\varepsilon}(\mathcal{P},\mathcal{P})
  - \mathcal{W}_{\varepsilon}(\mathcal{Q},\mathcal{Q}),
\label{eq:sinkhorn}
\end{equation}
where $\tilde{\mathcal{W}}_{\varepsilon}(\mathcal{P},\mathcal{Q})$ is called
Sinkhorn loss in Genevary et~al.~\cite{genevay2018learning}.

\subsubsection{Style Matching Loss}
We adopt Sinkhorn loss in Equation~\ref{eq:sinkhorn} to match the style
distributions of the source and target domains, so the style matching loss can be
expressed as
\begin{equation}
\mathcal{L}_s = 
  \sum_{l=1}^{L_f}{
    \tilde{\mathcal{W}}_{\varepsilon}(p^l_{\mu},   q^l_{\mu}) + 
    \tilde{\mathcal{W}}_{\varepsilon}(p^l_{\sigma},q^l_{\sigma}),
  }
\label{eq:stle-match}
\end{equation}
where $L_f$ is the number of adaptation layers in $f(x)$.


\subsection{Style-based Adaptation}

Since style matching loss in Equation~\ref{eq:stle-match} requires the supervision
of classification loss to learn categorization, adaptation loss with
style matching is defined as
\begin{equation}
\mathcal{L} = \mathcal{L}_c + \lambda\mathcal{L}_s
\end{equation}
where $\lambda$ is a Lagrange multiplier. In the experiments, to prevent
$\mathcal{L}_s$ from dominating the total loss, we set $\lambda = 0.01$ to make
the product of $\lambda\mathcal{L}_s$ less than $\mathcal{L}_c$.
The complete loss proposed in this paper consists of both PS
and SM:
\begin{equation}
\mathcal{L}^* = \mathcal{L}^*_c + \lambda\mathcal{L}_s.
\end{equation}
By minimizing $\mathcal{L}^*$, we adapt a model from the source domain to the
target domain from a style perspective.



\section{Evaluations}
\label{evaluations}

The two major applications of face recognition are face verification and identification,
of which face identification further divides into closed-set and
open-set identification. The performance of a model across different scenarios
must be evaluated under different protocols.

\subsection{Verification}

The task of verification is to judge whether the given two facial images are the
same identity. The two major evaluation protocols are described below.

\subsubsection{K-fold Cross Validation}
In this protocol, to avoid bias, the numbers of positive and negative pairs are usually the
same, such as in the Labeled Faces in the Wild (LFW)~\cite{LFWTech,
LFWTechUpdate} and YouTube Faces (YTF)~\cite{wolf2011face} datasets. The
dataset is divided into $K$ partitions. We evaluate the model on each partition by
using the optimal threshold found by testing all the other partitions, and
average the accuracies. 

\subsubsection{True Positive Rate under False Positive Rate}
We determine the threshold for rejecting negative pairs 
that corresponds to the desired false positive rate (FPR) or false alarm rate (FAR), 
which yields a true positive rate (TPR) that represents the accuracy on positive pairs.
In general, there are far more negative pairs than positive pairs to test the robustness of a model. 


\subsection{Closed-set Identification}

Given a facial image, usually called the probe, the goal of face identification is
to determine its identity by finding the most similar facial images in a
gallery. In the case of closed-set identification, the identities of the probes must
be a subset of the identities in the gallery. Therefore, rank-K (or top-K)
accuracy is employed to evaluate the performance under this scenario. For each
probe, we compute the similarities of all images in the gallery with respect to the
probe, and sort these similarities from high to low. If the similarity between the
probe and images of the same identity is ranked in the top-K, this identity is taken to
be the correct prediction. After testing all the probes, rank-K accuracy is
obtained as the ratio of correct predictions.


\subsection{Open-set Identification}

In open-set identification, the identities of probes may not be found in the gallery,
in which case the task is not only to determine the identity of a given
probe but also to report failure if the identity of the probe is not enrolled in the
gallery. To evaluate the performance of open-set identification, we first 
set a threshold by which to reject probes with unknown identities under an acceptable FAR; 
in this case, this is the false positive identification rate (FPIR).
Based on the set threshold, our metric is rank-1 accuracy, that is, the true
positive identification rate (TPIR). 



\section{Experiments}
\label{experiments}

\subsection{Datasets}

The datasets utilized here are listed in Table~\ref{tab:dataset}.
Created for different purposes, they can be divided into three types:

\subsubsection{Training}
VGGFace2~\cite{cao2018vggface2} and CASIA-WebFace~\cite{yi2014learning} are our
training datasets; all facial images in both datasets were captured in the
wild. Due to the large number of images for each identity in 
VGGFace2, it ensures better learning than CASIA-WebFace, even though 
it contains fewer identities.
As in most of the relevant literature~\cite{sohn2017unsupervised, luo2018deep, wang2020deep, wang2019racial},
CASIA-WebFace is a primary source dataset in this paper.
Following Wang et~al.~\cite{wang2020deep} and Arachchilage and 
Izquierdo~\cite{arachchilage2020clusterface}, we initialize the backbone model on VGGFace2 
after which we fine-tune the parameters on CASIA-WebFace.

\subsubsection{Validation}
We use LFW~\cite{LFWTech}, CFP-FP~\cite{sengupta2016frontal}, and 
AgeDB-30~\cite{moschoglou2017agedb} as the validation datasets; for each we use a
10-fold cross validation protocol. Their small size facilitates efficient model evaluations.
When training a baseline model, we test it on these datasets
after every epoch. After training, we pick the model whose average accuracy
is the highest of all the generations as the baseline model.

\subsubsection{Testing}
\begin{figure}[!ht]
\centering
\includegraphics[width=3.0in]{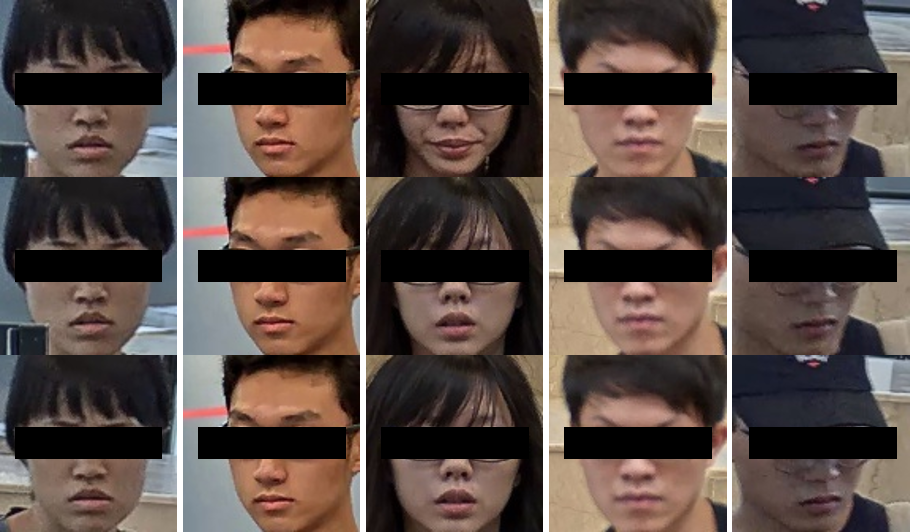}
\caption{Samples in NCTU-Lib dataset. 
Due to personal information protection, their eyes are masked.}
\label{fig:nctulib}
\end{figure}

IJB-A~\cite{klare2015pushing} is a joint face detection and recognition
dataset that contains a mixture of still images and videos under challenging
conditions, including blur as well as large pose and illumination variation. It provides
10-split evaluation protocols containing both verification and identification.
We take the average of the 10-split evaluations as the final results. 
IJB-A was extended as IJB-B~\cite{whitelam2017iarpa}, which was further extended as
IJB-C~\cite{maze2018iarpa}. The evaluation methods in IJB-B and IJB-C are the same. In terms
of verification, only one protocol is provided, but the positive-versus-negative 
pairs in IJB-B and IJB-C are about 10K/8M and 19.6K/15.6M respectively, which is more challenging
than that of IJB-A (1.8K/10K), as listed in Table~\ref{tab:protocol}.
For identification, both provide one probe set and two gallery sets for evaluation. 
We successively test each gallery set using the probe set and use their average as the final result. 
YTF~\cite{wolf2011face} contains 3,425 videos from 1,595 identities, for a total of 621.2K frames. 
As with the validation datasets, its protocol follows 10-fold cross validation, 
in which each fold is composed of 250 positive and negative pairs. 
Furthermore, we cooperate with the library of National Chiao Tung University (NCTU) 
for surveillance application. A system was built to collect the data automatically. 
When a subject entered the library, 
the system recorded and labeled the video by the student or staff identity card.
Due to personal information protection, the labels are anonymous.
To eliminate the interference of head pose variation, 
we adopted the head pose estimation model in Hsu et~al.~\cite{hsu2018quatnet}
to filter out the frames with large head pose (pitch, roll, or yaw $>15^\circ$).
This dataset is called NCTU-Lib, and its samples are shown in Fig.~\ref{fig:nctulib}.
The domain discrepancy between the training dataset and NCTU-Lib is large 
since almost all subjects in NCTU-Lib are Asian, which makes it more difficult to adapt the model.

\begin{table}[!ht]
\caption{Training and testing datasets. \\ 
Ver=verification, ID=identification.}
\label{tab:dataset}
\centering
\begin{adjustbox}{max width=\linewidth}
\begin{tabular}{lccccc}  
\toprule
\multirow{2}{*}{\bf Dataset} & \multicolumn{2}{c}{\bf Protocol} & \multirow{2}{*}{\bf Identities} & \multirow{2}{*}{\bf Images} & \multirow{2}{*}{\bf Images/identity} \\
\cline{2-3}				  & Ver & ID & & & \\
\hline
\hline
VGGFace2 \cite{cao2018vggface2} 	& & &  9,131 &   3.31M & 362.6 \\
CASIA-WebFace \cite{yi2014learning}	& & & 10,575 &  494.4K & 46.8 \\
\hline
LFW \cite{LFWTech}					& \checkmark & & 5,749 & 13,233 & 2.3 \\
CFP-FP \cite{sengupta2016frontal}	& \checkmark & &   500 &  7,000 &  14 \\
AgeDB-30 \cite{moschoglou2017agedb} & \checkmark & &   568 & 16,488 & 29 \\
\hline
IJB-A \cite{klare2015pushing}		& \checkmark & \checkmark &   500 &  25.8K &  51.6 \\
IJB-B \cite{whitelam2017iarpa}		& \checkmark & \checkmark & 1,845 &  76.8K &  41.6 \\
IJB-C \cite{maze2018iarpa}			& \checkmark & \checkmark & 3,531 & 148.8K &  42.1 \\
YTF \cite{wolf2011face} 			& \checkmark & 			  & 1,595 & 621.2K & 389.5 \\
\hline
NCTU-Lib							& \checkmark & \checkmark & 6,187 & 105.1K & 17.0 \\

\bottomrule
\end{tabular}
\end{adjustbox}
\end{table}

\begin{table}[!ht]
\caption{Evaluation protocols on the testing datasets.}
\label{tab:protocol}
\centering
\begin{adjustbox}{max width=\linewidth}
\begin{tabular}{lcccccc}  
\toprule
\multirow{3}{*}{\bf Dataset} & \multicolumn{2}{c}{\bf Verification} && \multicolumn{3}{c}{\bf Identification} \\
\cline{2-3}\cline{5-7}       & \multirow{2}{*}{Positive Pairs} & \multirow{2}{*}{Negative Pairs} && \multirow{2}{*}{Subjects (Gallery)} & \multicolumn{2}{c}{Samples/Subjects (Probe)} \\
\cline{6-7} & & && & Known & Unknown \\
\hline
\hline
YTF \cite{wolf2011face} 			&   250 &   250 &&    - &          - & \\
\hline
IJB-A \cite{klare2015pushing}		&  1.8K &   10K &&  112 &   1.2K/112 & 0.6K/55 \\
IJB-B \cite{whitelam2017iarpa}		&   10K &    8M &&  922 &   5.1K/922 & 5.1K/922 \\
IJB-C \cite{maze2018iarpa}			& 19.6K & 15.6M && 1.77K &  9.8K/1.77K & 9.8K/1.8K \\
\hline
NCTU-Lib							&  4.4K &  3.2M &&  3.16K &  6.0K/3.16K & 3.0K/3.0K \\
\bottomrule
\end{tabular}
\end{adjustbox}
\end{table}

\subsection{Implementation Details}

Here, we describe the image preprocessing, feature processing, baseline model training, source and
target dataset definitions, and training using the proposed domain adaptation methods.

\subsubsection{Processing of Images and Features}
If the multitask convolutional neural network (MTCNN)~\cite{zhang2016joint} 
detects a face in an image, we align the face to the fixed reference points
using the eyes, mouth corners, and nose center. If it
fails, we decide what to do based on the dataset. We ignore the training 
datasets and YTF since there is plenty of data for training and
testing. For IJB-A/B/C, if three landmarks (eyes and nose center) are provided, we
align these to the same reference points of the same landmarks, or we apply a
square box whose side is half of the maximal side of the image to crop on the
center. All processed images are resized to $128 \times 128$ and cropped to
$112 \times 112$ for testing. When training, we augment the images by mirroring
them with $50\%$ probability, and resize them to $128 \times 128$ and randomly
crop them to $112 \times 112$. In evaluations, we follow Whitelam et~al.~\cite{whitelam2017iarpa}
when fusing the features. 
That is, in a template, 
the frames of each type of media (still images or video frames) are averaged 
before all medias are averaged.

\subsubsection{Baseline Model and Source Dataset}
We used MobileFaceNet~\cite{chen2018mobilefacenets} as our backbone model.
Equation~\ref{eq:Lc} is the primary classification loss in this paper. As
mentioned above, the model was first trained on VGGFace2 dataset~\cite{cao2018vggface2}. 
We used stochastic gradient descent (SGD) with momentum to train all models. 
The training epochs, batch size, learning rate, and momentum were set to 50, 128, 0.1, and 0.9 respectively, and
the learning rate was divided by 10 every 12 epochs. 
After that, when transferring to CASIA-WebFace dataset~\cite{yi2014learning},
the learning rate was set to 0.01 to warm up the classifier $\hat{y}$ by one epoch 
before fine-tuning all parameters. 
We then set the learning rate to 0.0001 to fine-tune the parameters, 
and again divided by 10 every 12 epochs. The other settings remained unchanged.
The backbone model fine-tuned on CASIA-WebFace was then used as the baseline model denoted by softmax.

\subsubsection{Target Dataset}
We assumed that the IJB-A/B/C datasets~\cite{klare2015pushing, whitelam2017iarpa,
maze2018iarpa} are shared domain information. 
Only IJB-A was used as the target dataset when testing on these datasets. 
We used all images and frames in IJB-A to adapt the model. 
Similarly, when testing on the YTF dataset~\cite{wolf2011face},
we used all the frames of each video in YTF. 
During adaptation,
the learning rate was also set to 0.0001 to fine-tune the baseline model; 
other settings remained unchanged. 

\subsubsection{Regularizing Entropy for Style Matching}
Since the role of $\varepsilon$ is similar to bandwidth in MMD, we followed related 
studies~\cite{gretton2012optimal, luo2018deep, wang2020deep} to set its value. 
Luo et~al.~\cite{luo2018deep} and Wang et~al.~\cite{wang2020deep} suggest 
setting the bandwidth to the median pairwise distance on source dataset. 
However, such an exhaustive search is not practical since training datasets are usually large. 
This can instead be estimated by a Monte Carlo method, but we propose a more efficient way to determine the value.
Similar to the learning process in batch normalization, 
we estimated $\varepsilon$ by the mean pairwise distance between source and target samples in a mini-batch, 
and updated it every iteration by
\begin{equation}
\varepsilon_{k+1} = \rho\varepsilon_{k} + (1-\rho)\hat{\varepsilon}_{k},
\end{equation}
where $\hat{\varepsilon}_{k}$ is the current estimation. Momentum~$\rho$ was
set to $0.9$. 
In the following experiments, 
we compare the dynamic $\varepsilon$ (the proposed method) and the fixed $\varepsilon$.

\subsubsection{Definition of Adaptation Layers}
MobileFaceNet~\cite{chen2018mobilefacenets} is described in Table~\ref{tab:mfn}. 
For the following experiments, we mainly used four adaptation layers in shallow networks for SM. 
From shallow to deep, the dimensions of the feature maps were 
$56^2 \times 64$ ($l=1$), 
$28^2 \times 64$ ($l=2$), 
$28^2 \times 64$ ($l=3$), and 
$14^2 \times 128$  ($l=4$) respectively. 

\begin{table}[!ht]
\caption{MobileFaceNet architecture \cite{chen2018mobilefacenets} and proposed
layer index definitions for adaptation}
\label{tab:mfn}
\centering
\begin{adjustbox}{max width=\linewidth}
\begin{tabular}{lccc}  
\toprule
\bf Operator & \bf Input & \bf Output & $l$ \\
\hline
\hline
Convolution $3 \times 3$ 						&	$112^2 \times   3$ & 	$56^2 \times  64$ &	 - \\
Depth-wise convolution $3 \times 3$ 			&	$ 56^2 \times  64$ & 	$56^2 \times  64$ &	 1 \\
Bottleneck $\times 1 $							&	$ 56^2 \times  64$ & 	$28^2 \times  64$ &	 2 \\
Bottleneck $\times 4 $							&	$ 28^2 \times  64$ & 	$28^2 \times  64$ &	 3 \\
Bottleneck $\times 1 $							&	$ 28^2 \times  64$ & 	$14^2 \times 128$ &	 4 \\
Bottleneck $\times 6 $							&	$ 14^2 \times 128$ & 	$14^2 \times 128$ &	 5 \\
Bottleneck $\times 1 $							&	$ 14^2 \times 128$ & 	$ 7^2 \times 128$ &	 6 \\
Bottleneck $\times 2 $							&	$  7^2 \times 128$ & 	$ 7^2 \times 128$ &	 7 \\
Convolution $1 \times 1$ 						&	$  7^2 \times 128$ & 	$ 7^2 \times 512$ &	 8 \\
Linear depth-wise convolution $7 \times 7$ 		&	$  7^2 \times 512$ & 	$ 1^2 \times 512$ &	 9 \\
Linear convolution $1 \times 1$ 				&	$  1^2 \times 512$ & 	$ 1^2 \times 128$ &	10 \\
\bottomrule
\end{tabular}
\end{adjustbox}
\end{table}


\subsection{Ablation Study}

\begin{table*}[t]
\begin{minipage}{\textwidth}
\caption{Verification performance on IJB-A/B/C~\cite{klare2015pushing, whitelam2017iarpa, maze2018iarpa} for ablation experiments. \\
The values in boldface are the best in the row, and underlined values in boldface are the best in the column.}
\label{tab:ablation_ver}
\centering
\begin{adjustbox}{max width=\linewidth}
\begin{tabular}{lcccccccccccccc}  
\toprule
\multirow{2}{*}{\bf Method} & \multicolumn{4}{c}{\bf IJB-A TPR (\%)} && \multicolumn{4}{c}{\bf IJB-B TPR (\%)} && \multicolumn{4}{c}{\bf IJB-C TPR (\%)} \\
\cline{2-5}\cline{7-10}\cline{12-15} & FPR=0.0001 & FPR=0.001 & FPR=0.01 & FPR=0.1 && FPR=0.0001 & FPR=0.001 & FPR=0.01 & FPR=0.1 && FPR=0.0001 & FPR=0.001 & FPR=0.01 & FPR=0.1 \\
\hline
\hline

\bf Baseline & & & & && & & & && & & & \\
softmax				& 52.23 & 75.63 & 90.54 & 96.88 && 68.91 & 83.61 & 93.43 & 98.22 && 74.04 & 86.44 & 94.59 & 98.54 \\
\hline

\bf Perceptual scoring & & & & && & & & && & & & \\
softmax FF PS				& 54.60 & 77.67 & 91.02 & 96.88 && 70.29 & 84.15 & 93.74 & \bf98.32 && 75.02 & 87.00 & \bf94.92 & \bf98.61 \\
softmax PS					& \bf55.00 & \bf78.43 & \bf91.17 & \bf96.92 && \bf71.16 & \bf84.58 & \bf93.75 & 98.27 && \bf75.82 & \bf87.21 & 94.90 & 98.58 \\
\hline

\bf Style matching & & & & && & & & && & & & \\
softmax SM $(L_f=1$, fixed $\varepsilon)$ & 65.64 & 82.72	& 92.46 & \bf96.96 && 74.65 & 86.69 & 95.02 & 98.42 && 78.68 & 89.17 & 95.90 & 98.75 \\
softmax SM $(L_f=1)$					& 64.83 & 82.28 & 92.23 & \bf96.96 && 75.09 & 86.86 & 94.99 & 98.45 && 78.98 & 89.30 & 95.86 & 98.72 \\
softmax SM $(L_f=2)$					& 65.81 & \bf83.67 & \bf92.48 & 96.92 && 74.83 & 86.97 & \bf95.08 & \bf98.49 && 79.15 & 89.50 & \bf95.94 & \bf98.76 \\
softmax SM $(L_f=3)$					& 66.33 & 82.49 & 92.06 & 96.91 && \bf75.58 & 87.08 & 94.89 & 98.38 && 79.46 & 89.51 & 95.80 & 98.71 \\
softmax SM $(L_f=4)$					& \bf67.64 & 83.28 & 92.23 & 96.89 && 74.95 & \bf87.22 & 94.93 & 98.43 && \bf79.81 & \bf89.58 & 95.83 & 98.75 \\
\hline

\bf Combination & & & & && & & & && & & & \\
softmax SM $(L_f=1)$ + PS				& 66.88 & 83.53 & 92.48 & \bl{97.05} && 75.24 & 87.06 & \bl{95.15} & 98.47 && 79.76 & 89.56 & \bl{96.07} & 98.78 \\
softmax SM $(L_f=2)$ + PS				& 67.78 & \bl{84.31} & \bl{92.65} & 97.03 && \bl{75.96} & \bl{87.55} & 95.08 & \bl{98.50} && \bl{80.19} & \bl{89.85} & 95.99 & \bl{98.79} \\
softmax SM $(L_f=3)$ + PS				& \bl{68.70} & 83.52 & 92.30 & 96.82 && 75.51 & 87.32 & 95.00 & 98.48 && 79.83 & 89.83 & 95.92 & 98.78 \\
softmax SM $(L_f=4)$ + PS				& 68.58 & 83.76	& 92.31 & 96.77 && 75.50 & 87.20 & 94.93 & 98.48 && 79.93 & 89.61 & 95.81 & 98.71 \\

\bottomrule
\end{tabular}
\end{adjustbox} 
\end{minipage}
\end{table*}

\begin{table*}[t]
\begin{minipage}{\textwidth}
\caption{Identification performance on IJB-A/B/C~\cite{klare2015pushing, whitelam2017iarpa, maze2018iarpa} for ablation experiments. \\
The values in boldface are the best in the row, and underlined values in boldface are the best in the column.}
\label{tab:ablation_id}
\centering
\begin{adjustbox}{max width=\linewidth}
\begin{tabular}{lcccccccccccccc}  
\toprule
\multirow{2}{*}{\bf Method} & \multicolumn{4}{c}{\bf IJB-A TPIR (\%)} && \multicolumn{4}{c}{\bf IJB-B TPIR (\%)} && \multicolumn{4}{c}{\bf IJB-C TPIR (\%)} \\
\cline{2-5}\cline{7-10}\cline{12-15} & FPIR=0.01 & FPIR=0.1 & Rank-1 & Rank-10 && FPIR=0.01 & FPIR=0.1 & Rank-1 & Rank-10 && FPIR=0.01 & FPIR=0.1 & Rank-1 & Rank-10\\
\hline
\hline

\bf Baseline & & & & && & & & && & & & \\
softmax				& 65.31 & 85.74 & 94.79 & 97.91 && 59.77 & 77.10 & 88.01 & 95.22 && 58.86 & 76.49 & 88.93 & 95.07 \\
\hline

\bf Perceptual scoring & & & & && & & & && & & & \\
softmax FF PS				& 67.44	& 86.80 & \bf95.28 & \bf98.02 && \bf61.79 & 77.61 & \bf89.54 & \bf95.29 && 61.34 & \bf77.94 & 88.48 & \bf95.32 \\
softmax PS					& \bf68.78 & \bf86.95 & 95.09 & 98.00 && 61.76 & \bf77.98 & 88.42 & 95.28 && \bf62.18 & 77.76 & \bf89.35 & 95.29 \\
\hline

\bf Style matching & & & & && & & & && & & & \\
softmax SM $(L_f=1$, fixed $\varepsilon)$ & 76.28 & \bf89.35 & 95.13 & \bf98.11 && 64.44 & 80.42 & 88.29 & 95.61 && 64.90 & 80.32 & 89.68 & 95.55 \\
softmax SM $(L_f=1)$					& 76.14 & 89.06 & 95.17 & 98.02 && 65.85 & 79.98 & 88.34 & 95.50 && 66.07 & 80.26 & 89.71 & 95.60 \\
softmax SM $(L_f=2)$					& \bf76.59 & 89.14 & 95.10 & 98.03 && 64.77 & 80.43 & 88.42 & 95.56 && 66.68 & 80.68 & 89.74 & \bf95.61 \\
softmax SM $(L_f=3)$					& 76.06 & 89.05 & \bf95.20 & 98.08 && \bl{66.25} & 80.09 & 88.38 & \bf95.66 && \bf67.55 & 80.49 & 89.74 & 95.57 \\
softmax SM $(L_f=4)$					& 75.95 & 89.07 & 95.05 & 98.09 && 65.69 & \bf80.63 & \bf88.77 & 95.56 && 67.50 & \bf80.73 & \bf89.75 & 95.51 \\
\hline

\bf Combination & & & & && & & & && & & & \\
softmax SM $(L_f=1)$ + PS				& 77.35 & 89.43 & \bl{95.32} & 98.05 && 65.49 & 80.52 & 88.77 & 95.67 && 66.64 & 80.81 & \bl{89.98} & 95.62 \\
softmax SM $(L_f=2)$ + PS				& \bl{77.62} & \bl{89.65} & 95.15 & \bl{98.16} && 65.85 & \bl{80.87} & \bl{88.85} & 95.73 && 68.05 & \bl{81.20} & \bl{89.98} & \bl{95.74} \\
softmax SM $(L_f=3)$ + PS				& 77.35 & 89.06 & 95.10 & 98.00 && \bf65.94 & 80.72 & 88.61 & \bl{95.75} && \bl{68.22} & 80.94 & 89.90 & 95.73 \\
softmax SM $(L_f=4)$ + PS				& 76.83 & 89.02 & 95.16 & 98.02 && 65.30 & 80.50 & 88.80 & 95.66 && 68.07 & 80.98 & 89.89 & 95.65 \\

\bottomrule
\end{tabular}
\end{adjustbox} 
\end{minipage}
\end{table*}

\begin{table}[!ht]
\caption{Statistics for embeddings in IJB-A \cite{klare2015pushing}}
\label{tab:ablation_emb}
\centering
\begin{adjustbox}{max width=\linewidth}
\begin{tabular}{lcccc}  
\toprule
\multirow{2}{*}{\bf Method} & \multicolumn{2}{c}{\bf Cosine similarity} && \multirow{2}{*}{\bf Norm} \\
\cline{2-3} & Intra-class & Inter-class && \\
\hline
\hline
softmax									& 0.7074 $\pm$ 0.1325 & 0.0728 $\pm$ 0.1161 && 113.89 $\pm$ 14.30 \\
\hline
softmax PS								& 0.7071 $\pm$ 0.1330 & 0.0703 $\pm$ 0.1151 && 115.93 $\pm$ 13.76 \\
softmax SM $(L_f=2)$					& 0.6824 $\pm$ 0.1451 & 0.0191 $\pm$ 0.1103 && 118.21 $\pm$ 14.57 \\
softmax SM $(L_f=2)$ + PS				& 0.6836 $\pm$ 0.1445 & 0.0188 $\pm$ 0.1101 && 119.09 $\pm$ 14.17 \\
\bottomrule
\end{tabular}
\end{adjustbox} 
\end{table}

We analyzed our approaches using IJB-A as the
target dataset, and evaluated the performance on IJB-A/B/C. 
The testing results are reported in
Tables \ref{tab:ablation_ver}~and~\ref{tab:ablation_id}.

\subsubsection{Effectiveness of Perceptual Scoring}
\begin{figure}[!ht]
\centering
\includegraphics[width=3.0in]{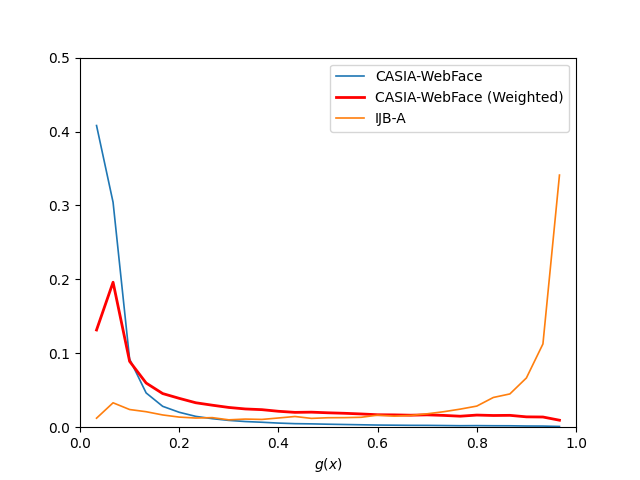}
\caption{Distributions of $g(x^s)$ and $g(x^t)$ under 
CASIA-WebFace~\cite{yi2014learning} and IJB-A~\cite{klare2015pushing}, and the weighted
distribution of CASIA-WebFace}
\label{fig:histo}
\end{figure}

The softmax PS method in Tables \ref{tab:ablation_ver}~and~\ref{tab:ablation_id} 
is the proposed PS method described in Equation~\ref{eq:wLc}. 
Compared with the baseline, it yields improved performance. 
We also adopted pure SqueezeNet~\cite{iandola2016squeezenet} as a discriminator, 
denoted as softmax FF PS, in which FF stands for feed-forward.
There is no significant difference between the two methods, but 
results show that softmax PS is robust to false alarms, 
which implies that low-level features are important for distinguishing identities at a finer granularity. 
To analyze the effect of PS, 
domain discriminator $g(x)$ is used to judge all images in both the source and target datasets. 
The distributions are shown in Fig.~\ref{fig:histo}: 
the two domains are clearly separable, which is evidence of domain discrepancy. 
The purpose of PS is to mitigate such discrepancy by scoring samples in the source domain. 
With Equation~\ref{eq:wLc}, the distribution of the source dataset is flatter: 
note the reduced discrepancy for CASIA-WebFace (Weighted) in Fig.~\ref{fig:histo}.

\subsubsection{Effectiveness of Style Matching}
Compared with PS, all SM approaches yield improvements in most metrics under different protocols. 
We find no differences among the metrics when varying the adaptation layers $L_f$, 
but we do note a trend in which TPR or TPIR under lower FAR is better 
when more adaptation layers are applied. 
However, there is no clear benefit to a fixed $\varepsilon$. 
It is slightly better for some metrics but worse on others. 
Conversely, using a dynamic $\varepsilon$ is more convenient, and yields similar performance.

\subsubsection{Combination of Perceptual Scoring and Style Matching}
The performance of SM is further improved by PS. 
Improvements are more obvious on verification protocols. 
The improvements on identification protocols are slight but consistent. 
That is,
most of the metrics show improvements. 
In particular, distinct improvements can be found in lower FAR (FPR and FPIR), 
which is an important concern in real-world scenarios. 
This implies that PS does help to generalize the learning of style matching on the target domain. 
Since most of the best metrics are achieved by $L_f=2$, we adopt this as our default setting.

\subsubsection{Embedding Analysis}
To understand how the proposed methods affect the embedding space, 
we measure the embedding norm and inter-class and intra-class cosine similarities on the target datasets, 
as listed in Table~\ref{tab:ablation_emb}. 
The norm measures the discriminative power of an embedding. 
PS slightly increases both the inter-class similarity and the embedding norm 
while preserving the intra-class similarity. 
SM separates embeddings from different classes by a large margin. 
Although intra-class similarity is decreased, 
the performance is also improved by larger inter-class similarity. 
In fact, the number of negative samples dominates in the testing protocols, 
which is a reasonable hypothesis in real-world applications; 
thus, slightly enlarging inter-class distances yields distinct improvements on accuracy. 
However, lower intra-class similarity saturates the performance at higher FAR, 
as shown in Tables \ref{tab:ablation_ver}~and~\ref{tab:ablation_id}. 
SM equipped with PS enjoys the benefits of both:
not only is inter-class separation further enlarged, but intra-class similarity is also preserved slightly.



\subsection{Benchmark Comparison}

\begin{table*}[t]
\begin{minipage}{\textwidth}
\caption{Benchmark comparison of verification performance on 
IJB-A/B/C~\cite{klare2015pushing, whitelam2017iarpa, maze2018iarpa}. \\
The values in boldface are the best under the softmax baseline, 
and underlined values in boldface are the best in the column.}
\label{tab:benchmark_ijb_ver}
\centering
\begin{adjustbox}{max width=\linewidth}
\begin{tabular}{lcccccccccccccc}  
\toprule
\multirow{2}{*}{\bf Method} & \multicolumn{4}{c}{\bf IJB-A TPR (\%)} && \multicolumn{4}{c}{\bf IJB-B TPR (\%)} && \multicolumn{4}{c}{\bf IJB-C TPR (\%)} \\
\cline{2-5}\cline{7-10}\cline{12-15} & FPR=0.0001 & FPR=0.001 & FPR=0.01 & FPR=0.1 && FPR=0.0001 & FPR=0.001 & FPR=0.01 & FPR=0.1 && FPR=0.0001 & FPR=0.001 & FPR=0.01 & FPR=0.1 \\
\hline
\hline
Sohn et al. \cite{sohn2017unsupervised}			& - & 58.40 & 82.80 & 96.20 && - & - & - & - && - & - & - & - \\
IMAN-A \cite{wang2019racial}					& - & \bl{84.49} & 91.88 & \bl{97.05} && - & - & - & - && - & - & - & - \\
CDA (vgg-soft) \cite{wang2020deep}				& - & 74.76 & 89.76 & 98.19 && - & - & - & - && - & - & - & - \\
CDA (res-arc) \cite{wang2020deep}				& - & 82.45 & 91.11 & 96.96 && - & 87.35 & 94.55 & 98.08 && - & 88.06 & 94.85 & 98.33 \\
\hline
softmax											& 52.23 & 75.63 & 90.54 & 96.88 && 68.91 & 83.61 & 93.43 & 98.22 && 74.04 & 86.44 & 94.59 & 98.54 \\
softmax BIN \cite{nam2018batch}					& 59.17 & 76.17 & 89.48 & 96.22 && 68.48 & 82.29 & 92.62 & 98.16 && 73.00 & 85.04 & 94.02 & 98.52 \\	
softmax $\mathit{MMD}_{m9,10}$ \cite{luo2018deep} 		& 65.64 & 83.50 & 92.35 & 97.02 && 75.00 & 86.83 & 94.88 & \bl{98.54} && 79.18 & 89.29 & 95.86 & \bl{98.84} \\
\hline
\bf softmax SM $(L_f=2)$ + PS					& \bl{67.78} & \bf{84.31} & \bl{92.65} & \bf{97.03} && \bl{75.96} & \bl{87.55} & \bl{95.08} & 98.50 && \bl{80.19} & \bl{89.85} & \bl{95.99} & 98.79 \\
\bottomrule
\end{tabular}
\end{adjustbox} 
\end{minipage}
\end{table*}

\begin{table*}[t]
\begin{minipage}{\textwidth}
\caption{Benchmark comparison of identification performance on IJB-A/B/C~\cite{klare2015pushing, whitelam2017iarpa, maze2018iarpa}. \\
The values in boldface are the best under the softmax baseline, 
and underlined values in boldface are the best in the column.}
\label{tab:benchmark_ijb_id}
\centering
\begin{adjustbox}{max width=\linewidth}
\begin{tabular}{lcccccccccccccc}  
\toprule
\multirow{2}{*}{\bf Method} & \multicolumn{4}{c}{\bf IJB-A TPIR (\%)} && \multicolumn{4}{c}{\bf IJB-B TPIR (\%)} && \multicolumn{4}{c}{\bf IJB-C TPIR (\%)} \\
\cline{2-5}\cline{7-10}\cline{12-15} & FPIR=0.01 & FPIR=0.1 & Rank-1 & Rank-10 && FPIR=0.01 & FPIR=0.1 & Rank-1 & Rank-10 && FPIR=0.01 & FPIR=0.1 & Rank-1 & Rank-10\\
\hline
\hline
Sohn et al. \cite{sohn2017unsupervised}			& - & - & 87.90 & 97.00 && - & - & - & - && - & - & - & - \\
IMAN-A \cite{wang2019racial}					& - & - & 94.05 & 98.04 && - & - & - & - && - & - & - & - \\
CDA (vgg-soft) \cite{wang2020deep}				& 66.85 & 85.32 & 94.89 & 99.23 && - & - & - & - && - & - & - & - \\
CDA (res-arc) \cite{wang2020deep}				& 75.49 & 87.76 & 93.61 & 97.62 && - & - & 86.22 & 93.33 && - & - & 88.19 & 93.70 \\
\hline
softmax											& 65.31 & 85.74 & 94.79 & 97.91 && 59.77 & 77.10 & 88.01 & 95.22 && 58.86 & 76.49 & 88.93 & 95.07 \\
softmax BIN \cite{nam2018batch}					& 67.91 & 84.17	& 94.33	& 97.92 && 61.29 & 75.10 & 86.33 & 94.63 && 61.43 & 75.13 & 87.51 & 94.68 \\
softmax $\mathit{MMD}_{m9,10}$ \cite{luo2018deep} 		& 76.40 & 89.21 & \bl{95.21} & 98.07 && 64.23 & 80.07 & 88.35 & 95.40 && 65.80 & 80.44 & 89.73 & 95.54 \\	
\hline
\bf softmax SM $(L_f=2)$ + PS					& \bl{77.62} & \bl{89.65} & 95.15 & \bl{98.16} && \bl{65.85} & \bl{80.87} & \bl{88.85} & \bl{95.73} && \bl{68.05} & \bl{81.20} & \bl{89.98} & \bl{95.74} \\
\bottomrule
\end{tabular}
\end{adjustbox} 
\end{minipage}
\end{table*}

To adapt our baseline model to ensure a fair comparison, 
we implemented batch-instance normalization (BIN)~\cite{nam2018batch} and 
multi-kernel MMD ($\mathit{MMD}_{m9,10}$)~\cite{luo2018deep}.
Subscripts $9$~and~$10$ indicate that 
the multi-kernel MMD is applied to layers $l=9$ and $l=10$, as listed in Table~\ref{tab:mfn}. 
Evaluation on the YTF~\cite{wolf2011face} dataset reveals that 
the proposed approach yields improvements over the baseline of more than $1\%$, 
which is the best of all methods under this benchmark. 
When BIN is applied on the baseline model, 
we find that it generalizes poorly. 
Comprehensive testing results on 
IJB-A/B/C~\cite{klare2015pushing, whitelam2017iarpa, maze2018iarpa} are listed in
Tables \ref{tab:benchmark_ijb_ver}~and~\ref{tab:benchmark_ijb_id}. 
We note that despite the weak capacity of our backbone 
model~\cite{chen2018mobilefacenets} (Wang et~al.~\cite{wang2020deep, wang2019racial} 
employ ResNet34~\cite{he2016deep} as their primary backbone), our approach 
outperforms prior work, even that of Wang et~al.~\cite{wang2020deep}, who guide
ResNet using ArcFace~\cite{deng2019arcface}. 
Since performance at higher FAR is nearly all saturated, 
metrics at lower FAR become critical to judge the model's performance. 
These results show that the proposed approach outperforms most prior work.

\begin{table}[!ht]
\caption{Benchmark comparison on YTF~\cite{wolf2011face}. \\
The value in boldface is the best in the column.}
\label{tab:benchmark_ytf}
\centering
\begin{adjustbox}{max width=3.5in}
\begin{tabular}{lc}  
\toprule
\bf Method & \bf Accuracy \\
\hline
\hline
Sohn et al. \cite{sohn2017unsupervised}		& 95.38 \\
\hline
softmax										& 94.46 \\
softmax BIN \cite{nam2018batch}				& 93.62 \\
softmax $\mathit{MMD}_{m9,10}$ \cite{luo2018deep}	& 95.28 \\
\hline
\bf softmax SM $(L_f=2)$ + PS				& \bf{95.52} \\
\bottomrule
\end{tabular}
\end{adjustbox}
\end{table}


\subsection{Application}

\begin{figure*}[!ht]

	\centering

	\subfigure[\# Subjects = 1000] {
		\centering
		\includegraphics[width=0.3\textwidth]{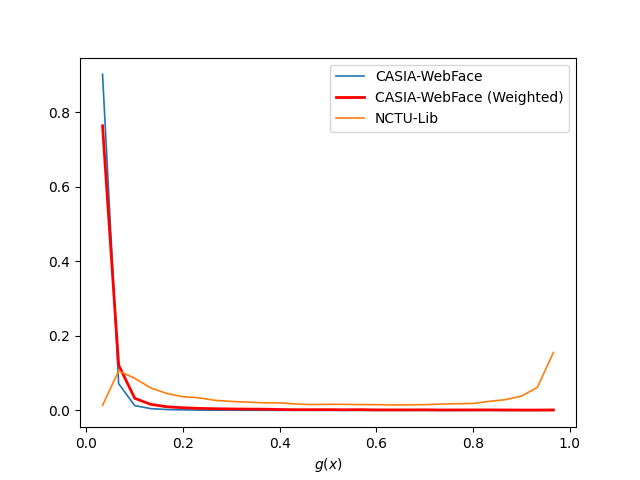}
		\label{fig:histo_nctu_1000}
	}
	\subfigure[\# Subjects = 3000] {
		\centering
		\includegraphics[width=0.3\textwidth]{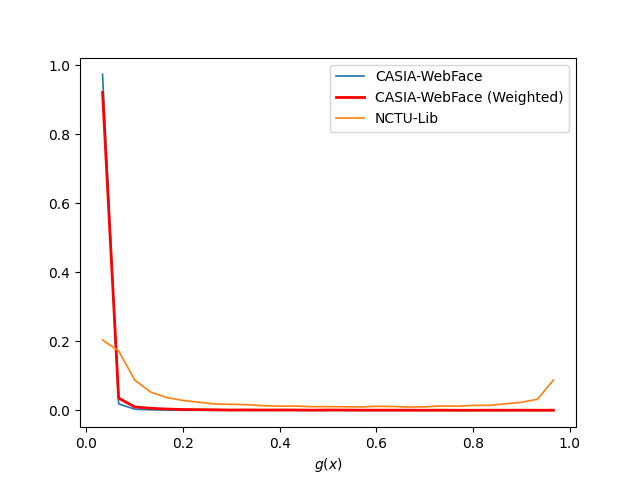}
		\label{fig:histo_nctu_3000}
	}
	\subfigure[\# Subjects = 5000] {
		\centering
		\includegraphics[width=0.3\textwidth]{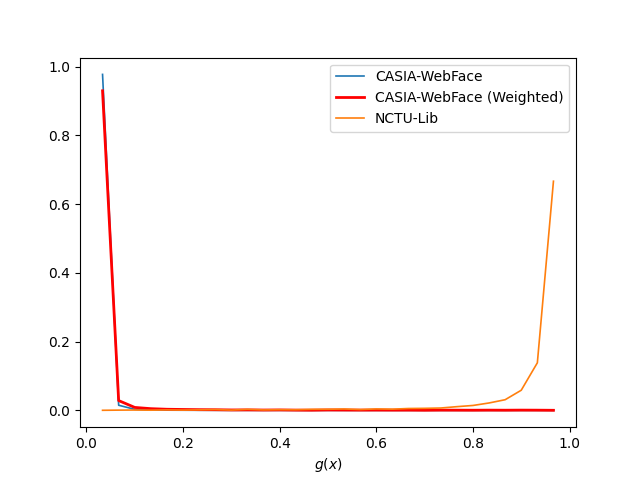}
		\label{fig:histo_nctu_5000}
	}

	\caption{Distributions of $g(x^s)$ and $g(x^t)$ under 
	CASIA-WebFace~\cite{yi2014learning} and NCTU-Lib, and the weighted distribution of CASIA-WebFace.}

	\label{fig:histo_nctu}

\end{figure*}

We apply our approaches to NCTU-Lib dataset here.
To simulate the real situation, we sampled 1,000, 3,000, and 5,000 subjects to be target datasets.
As we can see in Table~\ref{tab:app_nctulib}, no matter how many subjects are trained, 
the proposed approaches can improve the baseline model significantly, especially in lower FAR.
The major differences can be observed easily in the results of open-set identification under FPIR=0.01.
For softmax SM $(L_f=2)$, it is obvious that the improvements can be better with more subjects.
After applying PS, however, the performance decays with the growing of subjects. 
We speculate the reason is a lack of target domain information in the training dataset. 
As shown in Fig.~\ref{fig:histo_nctu}, with more subjects, 
it is easier for a discriminator $g(x)$ to distinguish two domains.
The scores of most source images are then close to 0, 
which results in down-weighting the classification loss.
Despite the drawback of PS, in Table~\ref{tab:app_nctulib}, softmax SM $(L_f=2)$ + PS with 1,000 subjects, 
since the weak discriminator can 
not only up-weight the target-like images but also preserve the weights of others, 
it does help SM to train a better model with fewer subjects.
Therefore, if the domain discrepancy is large, 
the discriminator cannot be too strong, 
or it may limit the performance.

\begin{table*}[t]
\begin{minipage}{\textwidth}
\caption{Verification and identification performance on NCTU-Lib. \\
The values in boldface are the best in the row, and underlined values in boldface are the best in the column.}
\label{tab:app_nctulib}
\centering
\begin{adjustbox}{max width=\linewidth}
\begin{tabular}{lccccccccccc}  
\toprule
\multirow{2}{*}{\bf Method} & \multirow{2}{*}{\bf \# Subjects} && \multicolumn{4}{c}{\bf Verification TPR (\%)} && \multicolumn{4}{c}{\bf Identification TPIR (\%)} \\
\cline{4-7}\cline{9-12} & && FPR=0.0001 & FPR=0.001 & FPR=0.01 & FPR=0.1 && FPIR=0.01 & FPIR=0.1 & Rank-1 & Rank-10 \\
\hline
\hline

softmax										&     - && 76.00 & 87.20 & 94.73 & 99.13 && 45.60 & 68.49 & 84.97 & 94.40 \\
\hline

\multirow{3}{*}{softmax SM $(L_f=2)$}		
											& 1,000 && 80.74 & \bf{90.39} & 96.64 & \bl{99.53} && 49.92 & \bf{74.27} & \bf{86.52} & 95.36 \\
											& 3,000 && 80.68 & \bf{90.39} & \bf{96.66} & \bl{99.53} && 50.77 & 73.64 & 86.12 & 95.36 \\
											& 5,000 && \bl{80.93} & 90.29 & 96.53 & 99.52 && \bf{51.48} & 73.64 & 86.40 & \bf{95.44} \\
\hline

\multirow{3}{*}{softmax SM $(L_f=2)$ + PS}	
											& 1,000 && \bf{80.67} & \bl{90.50} & \bl{96.84} & \bf{99.49} && \bl{51.74} & \bl{74.62} & \bl{86.59} & \bl{95.49} \\
											& 3,000 && 80.44 & 90.04 & 96.65 & 99.43 && 50.45 & 73.79 & 85.95 & 95.22 \\
											& 5,000 && 80.34 & 90.41 & 96.83 & 99.47 && 49.44 & 74.05 & 85.94 & 95.27 \\
\bottomrule
\end{tabular}
\end{adjustbox} 
\end{minipage}
\end{table*}

\section{Conclusion}
\label{conclusion}

In face recognition applications, classes do not overlap across domains; 
face recognition also involves fine-grained classification, 
and as such does not strictly follow the low-density separation principle. 
Together, these traits complicate the design of algorithms to reduce the domain gap. 
We treat domain mismatch as a style mismatch problem, 
and leverage texture-level features as the key to properly adapting a face recognition model to target domains. 
We propose perceptual scoring and style matching. 
The former mines visually target-like images in the training data to promote their learning, 
and the latter forces the learning of a common style space 
in which the style distributions of the two domains are confused, 
effectively normalizing the style of feature maps.
We conduct evaluations using the protocols of face verification and closed-set and open-set identification, 
demonstrating that our approach has good domain transfer ability and even outperforms most prior work. 
This progress is due to larger inter-class variation. 
The intra-class similarity, however, is reduced, 
which could explain the performance saturation at higher false alarm rates. 
Thus future work will involve reducing inter-class similarity and increasing intra-class similarity simultaneously 
to achieve comprehensive improvements on the target domain.


\section*{Declaration of Competing Interest}
The authors declare that they have no known competing financial interests 
or personal relationships that could have appeared to influence the work reported in this paper.

\section*{CRediT authorship contribution statement}

\textbf{Chun-Hsien~Lin:} Conceptuealization, Methodology, Software, Writing - Original Draft, Writing - Review \& Editing, Data Curation, Formal analysis, Investigation.
\newline
\textbf{Bing-Fei~Wu:} Supervision, Funding acquisition.

\section*{Acknowledgments}
This work was supported by the Ministry of Science and Technology
under Grant MOST 108-2638-E-009-001-MY2.

\bibliographystyle{elsarticle-num-names} 
\bibliography{ref}

\begin{thebibliography}{75}
\expandafter\ifx\csname natexlab\endcsname\relax\def\natexlab#1{#1}\fi
\providecommand{\url}[1]{\texttt{#1}}
\providecommand{\href}[2]{#2}
\providecommand{\path}[1]{#1}
\providecommand{\DOIprefix}{doi:}
\providecommand{\ArXivprefix}{arXiv:}
\providecommand{\URLprefix}{URL: }
\providecommand{\Pubmedprefix}{pmid:}
\providecommand{\doi}[1]{\href{http://dx.doi.org/#1}{\path{#1}}}
\providecommand{\Pubmed}[1]{\href{pmid:#1}{\path{#1}}}
\providecommand{\bibinfo}[2]{#2}
\ifx\xfnm\relax \def\xfnm[#1]{\unskip,\space#1}\fi
\bibitem[{Phillips(2017)}]{phillips2017cross}
\bibinfo{author}{P.~J. Phillips},
\newblock \bibinfo{title}{A cross benchmark assessment of a deep convolutional
  neural network for face recognition},
\newblock in: \bibinfo{booktitle}{12th IEEE International Conference on
  Automatic Face \& Gesture Recognition (FG 2017)},
  \bibinfo{organization}{IEEE}, \bibinfo{year}{2017}, pp.
  \bibinfo{pages}{705--710}.
\bibitem[{Luo et~al.(2018)Luo, Hu, Deng, and Shen}]{luo2018deep}
\bibinfo{author}{Z.~Luo}, \bibinfo{author}{J.~Hu}, \bibinfo{author}{W.~Deng},
  \bibinfo{author}{H.~Shen},
\newblock \bibinfo{title}{Deep unsupervised domain adaptation for face
  recognition},
\newblock in: \bibinfo{booktitle}{13th IEEE International Conference on
  Automatic Face \& Gesture Recognition (FG 2018)},
  \bibinfo{organization}{IEEE}, \bibinfo{year}{2018}, pp.
  \bibinfo{pages}{453--457}.
\bibitem[{Wang and Deng(2020)}]{wang2020deep}
\bibinfo{author}{M.~Wang}, \bibinfo{author}{W.~Deng},
\newblock \bibinfo{title}{Deep face recognition with clustering based domain
  adaptation},
\newblock \bibinfo{journal}{Neurocomputing} \bibinfo{volume}{393}
  (\bibinfo{year}{2020}) \bibinfo{pages}{1--14}.
\bibitem[{Wang et~al.(2019)Wang, Deng, Hu, Tao, and Huang}]{wang2019racial}
\bibinfo{author}{M.~Wang}, \bibinfo{author}{W.~Deng}, \bibinfo{author}{J.~Hu},
  \bibinfo{author}{X.~Tao}, \bibinfo{author}{Y.~Huang},
\newblock \bibinfo{title}{Racial faces in the wild: Reducing racial bias by
  information maximization adaptation network},
\newblock in: \bibinfo{booktitle}{Proceedings of the IEEE/CVF International
  Conference on Computer Vision}, \bibinfo{year}{2019}, pp.
  \bibinfo{pages}{692--702}.
\bibitem[{Arachchilage and Izquierdo(2020)}]{arachchilage2020ssdl}
\bibinfo{author}{S.~W. Arachchilage}, \bibinfo{author}{E.~Izquierdo},
\newblock \bibinfo{title}{{SSDL}: Self-supervised domain learning for improved
  face recognition},
\newblock \bibinfo{journal}{arXiv preprint arXiv:2011.13361}
  (\bibinfo{year}{2020}).
\bibitem[{Sohn et~al.(2017)Sohn, Liu, Zhong, Yu, Yang, and
  Chandraker}]{sohn2017unsupervised}
\bibinfo{author}{K.~Sohn}, \bibinfo{author}{S.~Liu},
  \bibinfo{author}{G.~Zhong}, \bibinfo{author}{X.~Yu}, \bibinfo{author}{M.-H.
  Yang}, \bibinfo{author}{M.~Chandraker},
\newblock \bibinfo{title}{Unsupervised domain adaptation for face recognition
  in unlabeled videos},
\newblock in: \bibinfo{booktitle}{Proceedings of the IEEE International
  Conference on Computer Vision}, \bibinfo{year}{2017}, pp.
  \bibinfo{pages}{3210--3218}.
\bibitem[{Hong et~al.(2017)Hong, Im, Ryu, and Yang}]{hong2017sspp}
\bibinfo{author}{S.~Hong}, \bibinfo{author}{W.~Im}, \bibinfo{author}{J.~Ryu},
  \bibinfo{author}{H.~S. Yang},
\newblock \bibinfo{title}{{SSPP-DAN}: Deep domain adaptation network for face
  recognition with single sample per person},
\newblock in: \bibinfo{booktitle}{2017 IEEE International Conference on Image
  Processing (ICIP)}, \bibinfo{organization}{IEEE}, \bibinfo{year}{2017}, pp.
  \bibinfo{pages}{825--829}.
\bibitem[{Arachchilage and Izquierdo(2020)}]{arachchilage2020clusterface}
\bibinfo{author}{S.~W. Arachchilage}, \bibinfo{author}{E.~Izquierdo},
\newblock \bibinfo{title}{{ClusterFace}: Joint clustering and classification
  for set-based face recognition},
\newblock \bibinfo{journal}{arXiv preprint arXiv:2011.13360}
  (\bibinfo{year}{2020}).
\bibitem[{Yosinski et~al.(2014)Yosinski, Clune, Bengio, and
  Lipson}]{yosinski2014transferable}
\bibinfo{author}{J.~Yosinski}, \bibinfo{author}{J.~Clune},
  \bibinfo{author}{Y.~Bengio}, \bibinfo{author}{H.~Lipson},
\newblock \bibinfo{title}{How transferable are features in deep neural
  networks?},
\newblock \bibinfo{journal}{arXiv preprint arXiv:1411.1792}
  (\bibinfo{year}{2014}).
\bibitem[{Bousmalis et~al.(2017)Bousmalis, Silberman, Dohan, Erhan, and
  Krishnan}]{bousmalis2017unsupervised}
\bibinfo{author}{K.~Bousmalis}, \bibinfo{author}{N.~Silberman},
  \bibinfo{author}{D.~Dohan}, \bibinfo{author}{D.~Erhan},
  \bibinfo{author}{D.~Krishnan},
\newblock \bibinfo{title}{Unsupervised pixel-level domain adaptation with
  generative adversarial networks},
\newblock in: \bibinfo{booktitle}{CVPR}, \bibinfo{year}{2017}.
\bibitem[{Tran et~al.(2019)Tran, Sohn, Yu, Liu, and Chandraker}]{tran2019gotta}
\bibinfo{author}{L.~Tran}, \bibinfo{author}{K.~Sohn}, \bibinfo{author}{X.~Yu},
  \bibinfo{author}{X.~Liu}, \bibinfo{author}{M.~Chandraker},
\newblock \bibinfo{title}{{Gotta Adapt 'Em All}: Joint pixel and feature-level
  domain adaptation for recognition in the wild},
\newblock in: \bibinfo{booktitle}{Proceedings of the IEEE/CVF Conference on
  Computer Vision and Pattern Recognition}, \bibinfo{year}{2019}, pp.
  \bibinfo{pages}{2672--2681}.
\bibitem[{Li et~al.(2020)Li, Jiao, Cao, Wong, and Wu}]{li2020model}
\bibinfo{author}{R.~Li}, \bibinfo{author}{Q.~Jiao}, \bibinfo{author}{W.~Cao},
  \bibinfo{author}{H.-S. Wong}, \bibinfo{author}{S.~Wu},
\newblock \bibinfo{title}{Model adaptation: Unsupervised domain adaptation
  without source data},
\newblock in: \bibinfo{booktitle}{Proceedings of the IEEE/CVF Conference on
  Computer Vision and Pattern Recognition}, \bibinfo{year}{2020}, pp.
  \bibinfo{pages}{9641--9650}.
\bibitem[{Goodfellow et~al.(2014)Goodfellow, Pouget-Abadie, Mirza, Xu,
  Warde-Farley, Ozair, Courville, and Bengio}]{goodfellow2014generative}
\bibinfo{author}{I.~J. Goodfellow}, \bibinfo{author}{J.~Pouget-Abadie},
  \bibinfo{author}{M.~Mirza}, \bibinfo{author}{B.~Xu},
  \bibinfo{author}{D.~Warde-Farley}, \bibinfo{author}{S.~Ozair},
  \bibinfo{author}{A.~Courville}, \bibinfo{author}{Y.~Bengio},
\newblock \bibinfo{title}{Generative adversarial networks},
\newblock \bibinfo{journal}{arXiv preprint arXiv:1406.2661}
  (\bibinfo{year}{2014}).
\bibitem[{Karras et~al.(2017)Karras, Aila, Laine, and
  Lehtinen}]{karras2017progressive}
\bibinfo{author}{T.~Karras}, \bibinfo{author}{T.~Aila},
  \bibinfo{author}{S.~Laine}, \bibinfo{author}{J.~Lehtinen},
\newblock \bibinfo{title}{Progressive growing of {GANs} for improved quality,
  stability, and variation},
\newblock \bibinfo{journal}{arXiv preprint arXiv:1710.10196}
  (\bibinfo{year}{2017}).
\bibitem[{Karras et~al.(2019)Karras, Laine, and Aila}]{karras2019style}
\bibinfo{author}{T.~Karras}, \bibinfo{author}{S.~Laine},
  \bibinfo{author}{T.~Aila},
\newblock \bibinfo{title}{A style-based generator architecture for generative
  adversarial networks},
\newblock in: \bibinfo{booktitle}{Proceedings of the IEEE/CVF Conference on
  Computer Vision and Pattern Recognition}, \bibinfo{year}{2019}, pp.
  \bibinfo{pages}{4401--4410}.
\bibitem[{Tewari et~al.(2020)Tewari, Elgharib, Bharaj, Bernard, Seidel,
  P{\'e}rez, Zollhofer, and Theobalt}]{tewari2020stylerig}
\bibinfo{author}{A.~Tewari}, \bibinfo{author}{M.~Elgharib},
  \bibinfo{author}{G.~Bharaj}, \bibinfo{author}{F.~Bernard},
  \bibinfo{author}{H.-P. Seidel}, \bibinfo{author}{P.~P{\'e}rez},
  \bibinfo{author}{M.~Zollhofer}, \bibinfo{author}{C.~Theobalt},
\newblock \bibinfo{title}{{StyleRig}: Rigging {StyleGAN} for {3D} control over
  portrait images},
\newblock in: \bibinfo{booktitle}{Proceedings of the IEEE/CVF Conference on
  Computer Vision and Pattern Recognition}, \bibinfo{year}{2020}, pp.
  \bibinfo{pages}{6142--6151}.
\bibitem[{Zhang et~al.(2018)Zhang, Isola, Efros, Shechtman, and
  Wang}]{zhang2018unreasonable}
\bibinfo{author}{R.~Zhang}, \bibinfo{author}{P.~Isola}, \bibinfo{author}{A.~A.
  Efros}, \bibinfo{author}{E.~Shechtman}, \bibinfo{author}{O.~Wang},
\newblock \bibinfo{title}{The unreasonable effectiveness of deep features as a
  perceptual metric},
\newblock in: \bibinfo{booktitle}{Proceedings of the IEEE Conference on
  Computer Vision and Pattern Recognition}, \bibinfo{year}{2018}, pp.
  \bibinfo{pages}{586--595}.
\bibitem[{Chu et~al.(2016)Chu, De~la Torre, and Cohn}]{chu2016selective}
\bibinfo{author}{W.-S. Chu}, \bibinfo{author}{F.~De~la Torre},
  \bibinfo{author}{J.~F. Cohn},
\newblock \bibinfo{title}{Selective transfer machine for personalized facial
  expression analysis},
\newblock \bibinfo{journal}{IEEE Transactions on Pattern Analysis and Machine
  Intelligence} \bibinfo{volume}{39} (\bibinfo{year}{2016})
  \bibinfo{pages}{529--545}.
\bibitem[{Wu and Lin(2018)}]{wu2018adaptive}
\bibinfo{author}{B.-F. Wu}, \bibinfo{author}{C.-H. Lin},
\newblock \bibinfo{title}{Adaptive feature mapping for customizing deep
  learning based facial expression recognition model},
\newblock \bibinfo{journal}{IEEE Access} \bibinfo{volume}{6}
  (\bibinfo{year}{2018}) \bibinfo{pages}{12451--12461}.
\bibitem[{Liu et~al.(2019)Liu, Song, Zou, and Zhang}]{liu2019reinforced}
\bibinfo{author}{M.~Liu}, \bibinfo{author}{Y.~Song}, \bibinfo{author}{H.~Zou},
  \bibinfo{author}{T.~Zhang},
\newblock \bibinfo{title}{Reinforced training data selection for domain
  adaptation},
\newblock in: \bibinfo{booktitle}{Proceedings of the 57th Annual Meeting of the
  Association for Computational Linguistics}, \bibinfo{year}{2019}, pp.
  \bibinfo{pages}{1957--1968}.
\bibitem[{Ulyanov et~al.(2017)Ulyanov, Vedaldi, and
  Lempitsky}]{ulyanov2017improved}
\bibinfo{author}{D.~Ulyanov}, \bibinfo{author}{A.~Vedaldi},
  \bibinfo{author}{V.~Lempitsky},
\newblock \bibinfo{title}{Improved texture networks: Maximizing quality and
  diversity in feed-forward stylization and texture synthesis},
\newblock in: \bibinfo{booktitle}{Proceedings of the IEEE Conference on
  Computer Vision and Pattern Recognition}, \bibinfo{year}{2017}, pp.
  \bibinfo{pages}{6924--6932}.
\bibitem[{Dumoulin et~al.(2016)Dumoulin, Shlens, and
  Kudlur}]{dumoulin2016learned}
\bibinfo{author}{V.~Dumoulin}, \bibinfo{author}{J.~Shlens},
  \bibinfo{author}{M.~Kudlur},
\newblock \bibinfo{title}{A learned representation for artistic style},
\newblock \bibinfo{journal}{arXiv preprint arXiv:1610.07629}
  (\bibinfo{year}{2016}).
\bibitem[{Huang and Belongie(2017)}]{huang2017arbitrary}
\bibinfo{author}{X.~Huang}, \bibinfo{author}{S.~Belongie},
\newblock \bibinfo{title}{Arbitrary style transfer in real-time with adaptive
  instance normalization},
\newblock in: \bibinfo{booktitle}{Proceedings of the IEEE International
  Conference on Computer Vision}, \bibinfo{year}{2017}, pp.
  \bibinfo{pages}{1501--1510}.
\bibitem[{Cuturi(2013)}]{cuturi2013sinkhorn}
\bibinfo{author}{M.~Cuturi},
\newblock \bibinfo{title}{Sinkhorn distances: lightspeed computation of optimal
  transport},
\newblock in: \bibinfo{booktitle}{NIPS}, volume~\bibinfo{volume}{2},
  \bibinfo{year}{2013}, p.~\bibinfo{pages}{4}.
\bibitem[{Cuturi and Doucet(2014)}]{cuturi2014fast}
\bibinfo{author}{M.~Cuturi}, \bibinfo{author}{A.~Doucet},
\newblock \bibinfo{title}{Fast computation of {W}asserstein barycenters},
\newblock in: \bibinfo{booktitle}{International Conference on Machine
  Learning}, \bibinfo{organization}{PMLR}, \bibinfo{year}{2014}, pp.
  \bibinfo{pages}{685--693}.
\bibitem[{Aude et~al.(2016)Aude, Cuturi, Peyr{\'e}, and
  Bach}]{aude2016stochastic}
\bibinfo{author}{G.~Aude}, \bibinfo{author}{M.~Cuturi},
  \bibinfo{author}{G.~Peyr{\'e}}, \bibinfo{author}{F.~Bach},
\newblock \bibinfo{title}{Stochastic optimization for large-scale optimal
  transport},
\newblock \bibinfo{journal}{arXiv preprint arXiv:1605.08527}
  (\bibinfo{year}{2016}).
\bibitem[{Feydy et~al.(2019)Feydy, S{\'e}journ{\'e}, Vialard, Amari,
  Trouv{\'e}, and Peyr{\'e}}]{feydy2019interpolating}
\bibinfo{author}{J.~Feydy}, \bibinfo{author}{T.~S{\'e}journ{\'e}},
  \bibinfo{author}{F.-X. Vialard}, \bibinfo{author}{S.-i. Amari},
  \bibinfo{author}{A.~Trouv{\'e}}, \bibinfo{author}{G.~Peyr{\'e}},
\newblock \bibinfo{title}{Interpolating between optimal transport and {MMD}
  using {S}inkhorn divergences},
\newblock in: \bibinfo{booktitle}{22nd International Conference on Artificial
  Intelligence and Statistics}, \bibinfo{organization}{PMLR},
  \bibinfo{year}{2019}, pp. \bibinfo{pages}{2681--2690}.
\bibitem[{Genevay et~al.(2019)Genevay, Chizat, Bach, Cuturi, and
  Peyr{\'e}}]{genevay2019sample}
\bibinfo{author}{A.~Genevay}, \bibinfo{author}{L.~Chizat},
  \bibinfo{author}{F.~Bach}, \bibinfo{author}{M.~Cuturi},
  \bibinfo{author}{G.~Peyr{\'e}},
\newblock \bibinfo{title}{Sample complexity of {S}inkhorn divergences},
\newblock in: \bibinfo{booktitle}{22nd International Conference on Artificial
  Intelligence and Statistics}, \bibinfo{organization}{PMLR},
  \bibinfo{year}{2019}, pp. \bibinfo{pages}{1574--1583}.
\bibitem[{Genevay(2019)}]{genevay2019entropy}
\bibinfo{author}{A.~Genevay}, \bibinfo{title}{Entropy-regularized optimal
  transport for machine learning}, Ph.D. thesis, Paris Sciences et Lettres,
  \bibinfo{year}{2019}.
\bibitem[{Chizat et~al.(2020)Chizat, Roussillon, L{\'e}ger, Vialard, and
  Peyr{\'e}}]{chizat2020faster}
\bibinfo{author}{L.~Chizat}, \bibinfo{author}{P.~Roussillon},
  \bibinfo{author}{F.~L{\'e}ger}, \bibinfo{author}{F.-X. Vialard},
  \bibinfo{author}{G.~Peyr{\'e}},
\newblock \bibinfo{title}{Faster {W}asserstein distance estimation with the
  {S}inkhorn divergence},
\newblock \bibinfo{journal}{arXiv preprint arXiv:2006.08172}
  (\bibinfo{year}{2020}).
\bibitem[{Klare et~al.(2015)Klare, Klein, Taborsky, Blanton, Cheney, Allen,
  Grother, Mah, and Jain}]{klare2015pushing}
\bibinfo{author}{B.~F. Klare}, \bibinfo{author}{B.~Klein},
  \bibinfo{author}{E.~Taborsky}, \bibinfo{author}{A.~Blanton},
  \bibinfo{author}{J.~Cheney}, \bibinfo{author}{K.~Allen},
  \bibinfo{author}{P.~Grother}, \bibinfo{author}{A.~Mah},
  \bibinfo{author}{A.~K. Jain},
\newblock \bibinfo{title}{Pushing the frontiers of unconstrained face detection
  and recognition: {IARPA Janus Benchmark A}},
\newblock in: \bibinfo{booktitle}{Proceedings of the IEEE Conference on
  Computer Vision and Pattern Recognition}, \bibinfo{year}{2015}, pp.
  \bibinfo{pages}{1931--1939}.
\bibitem[{Whitelam et~al.(2017)Whitelam, Taborsky, Blanton, Maze, Adams,
  Miller, Kalka, Jain, Duncan, Allen et~al.}]{whitelam2017iarpa}
\bibinfo{author}{C.~Whitelam}, \bibinfo{author}{E.~Taborsky},
  \bibinfo{author}{A.~Blanton}, \bibinfo{author}{B.~Maze},
  \bibinfo{author}{J.~Adams}, \bibinfo{author}{T.~Miller},
  \bibinfo{author}{N.~Kalka}, \bibinfo{author}{A.~K. Jain},
  \bibinfo{author}{J.~A. Duncan}, \bibinfo{author}{K.~Allen}, et~al.,
\newblock \bibinfo{title}{{IARPA Janus Benchmark-B} face dataset},
\newblock in: \bibinfo{booktitle}{Proceedings of the IEEE Conference on
  Computer Vision and Pattern Recognition Workshops}, \bibinfo{year}{2017}, pp.
  \bibinfo{pages}{90--98}.
\bibitem[{Maze et~al.(2018)Maze, Adams, Duncan, Kalka, Miller, Otto, Jain,
  Niggel, Anderson, Cheney et~al.}]{maze2018iarpa}
\bibinfo{author}{B.~Maze}, \bibinfo{author}{J.~Adams}, \bibinfo{author}{J.~A.
  Duncan}, \bibinfo{author}{N.~Kalka}, \bibinfo{author}{T.~Miller},
  \bibinfo{author}{C.~Otto}, \bibinfo{author}{A.~K. Jain},
  \bibinfo{author}{W.~T. Niggel}, \bibinfo{author}{J.~Anderson},
  \bibinfo{author}{J.~Cheney}, et~al.,
\newblock \bibinfo{title}{{IARPA Janus Benchmark-C}: Face dataset and
  protocol},
\newblock in: \bibinfo{booktitle}{International Conference on Biometrics
  (ICB)}, \bibinfo{organization}{IEEE}, \bibinfo{year}{2018}, pp.
  \bibinfo{pages}{158--165}.
\bibitem[{Taigman et~al.(2014)Taigman, Yang, Ranzato, and
  Wolf}]{taigman2014deepface}
\bibinfo{author}{Y.~Taigman}, \bibinfo{author}{M.~Yang},
  \bibinfo{author}{M.~Ranzato}, \bibinfo{author}{L.~Wolf},
\newblock \bibinfo{title}{{DeepFace}: Closing the gap to human-level
  performance in face verification},
\newblock in: \bibinfo{booktitle}{Proceedings of the IEEE Conference on
  Computer Vision and Pattern Recognition}, \bibinfo{year}{2014}, pp.
  \bibinfo{pages}{1701--1708}.
\bibitem[{Huang et~al.(2007)Huang, Ramesh, Berg, and Learned-Miller}]{LFWTech}
\bibinfo{author}{G.~B. Huang}, \bibinfo{author}{M.~Ramesh},
  \bibinfo{author}{T.~Berg}, \bibinfo{author}{E.~Learned-Miller},
  \bibinfo{title}{Labeled {F}aces in the {W}ild: A Database for Studying Face
  Recognition in Unconstrained Environments}, \bibinfo{type}{Technical Report}
  \bibinfo{number}{07-49}, University of Massachusetts, Amherst,
  \bibinfo{year}{2007}.
\bibitem[{Masi et~al.(2018)Masi, Wu, Hassner, and Natarajan}]{masi2018deep}
\bibinfo{author}{I.~Masi}, \bibinfo{author}{Y.~Wu},
  \bibinfo{author}{T.~Hassner}, \bibinfo{author}{P.~Natarajan},
\newblock \bibinfo{title}{Deep face recognition: A survey},
\newblock in: \bibinfo{booktitle}{31st Conference on Graphics, Patterns and
  Images (SIBGRAPI)}, \bibinfo{organization}{IEEE}, \bibinfo{year}{2018}, pp.
  \bibinfo{pages}{471--478}.
\bibitem[{Ouyang et~al.(2014)Ouyang, Luo, Zeng, Qiu, Tian, Li, Yang, Wang,
  Xiong, Qian et~al.}]{ouyang2014deepid}
\bibinfo{author}{W.~Ouyang}, \bibinfo{author}{P.~Luo},
  \bibinfo{author}{X.~Zeng}, \bibinfo{author}{S.~Qiu},
  \bibinfo{author}{Y.~Tian}, \bibinfo{author}{H.~Li},
  \bibinfo{author}{S.~Yang}, \bibinfo{author}{Z.~Wang},
  \bibinfo{author}{Y.~Xiong}, \bibinfo{author}{C.~Qian}, et~al.,
\newblock \bibinfo{title}{{DeepID-Net}: Multi-stage and deformable deep
  convolutional neural networks for object detection},
\newblock \bibinfo{journal}{arXiv preprint arXiv:1409.3505}
  (\bibinfo{year}{2014}).
\bibitem[{Ouyang et~al.(2015)Ouyang, Wang, Zeng, Qiu, Luo, Tian, Li, Yang,
  Wang, Loy et~al.}]{ouyang2015deepid}
\bibinfo{author}{W.~Ouyang}, \bibinfo{author}{X.~Wang},
  \bibinfo{author}{X.~Zeng}, \bibinfo{author}{S.~Qiu},
  \bibinfo{author}{P.~Luo}, \bibinfo{author}{Y.~Tian}, \bibinfo{author}{H.~Li},
  \bibinfo{author}{S.~Yang}, \bibinfo{author}{Z.~Wang}, \bibinfo{author}{C.-C.
  Loy}, et~al.,
\newblock \bibinfo{title}{{DeepID-Net}: Deformable deep convolutional neural
  networks for object detection},
\newblock in: \bibinfo{booktitle}{Proceedings of the IEEE Conference on
  Computer Vision and Pattern Recognition}, \bibinfo{year}{2015}, pp.
  \bibinfo{pages}{2403--2412}.
\bibitem[{Ouyang et~al.(2016)Ouyang, Zeng, Wang, Qiu, Luo, Tian, Li, Yang,
  Wang, Li et~al.}]{ouyang2016deepid}
\bibinfo{author}{W.~Ouyang}, \bibinfo{author}{X.~Zeng},
  \bibinfo{author}{X.~Wang}, \bibinfo{author}{S.~Qiu},
  \bibinfo{author}{P.~Luo}, \bibinfo{author}{Y.~Tian}, \bibinfo{author}{H.~Li},
  \bibinfo{author}{S.~Yang}, \bibinfo{author}{Z.~Wang},
  \bibinfo{author}{H.~Li}, et~al.,
\newblock \bibinfo{title}{{DeepID-Net}: Object detection with deformable part
  based convolutional neural networks},
\newblock \bibinfo{journal}{IEEE Transactions on Pattern Analysis and Machine
  Intelligence} \bibinfo{volume}{39} (\bibinfo{year}{2016})
  \bibinfo{pages}{1320--1334}.
\bibitem[{Hadsell et~al.(2006)Hadsell, Chopra, and
  LeCun}]{hadsell2006dimensionality}
\bibinfo{author}{R.~Hadsell}, \bibinfo{author}{S.~Chopra},
  \bibinfo{author}{Y.~LeCun},
\newblock \bibinfo{title}{Dimensionality reduction by learning an invariant
  mapping},
\newblock in: \bibinfo{booktitle}{IEEE Computer Society Conference on Computer
  Vision and Pattern Recognition (CVPR'06)}, volume~\bibinfo{volume}{2},
  \bibinfo{organization}{IEEE}, \bibinfo{year}{2006}, pp.
  \bibinfo{pages}{1735--1742}.
\bibitem[{Schroff et~al.(2015)Schroff, Kalenichenko, and
  Philbin}]{schroff2015facenet}
\bibinfo{author}{F.~Schroff}, \bibinfo{author}{D.~Kalenichenko},
  \bibinfo{author}{J.~Philbin},
\newblock \bibinfo{title}{{FaceNet}: A unified embedding for face recognition
  and clustering},
\newblock in: \bibinfo{booktitle}{Proceedings of the IEEE Conference on
  Computer Vision and Pattern Recognition}, \bibinfo{year}{2015}, pp.
  \bibinfo{pages}{815--823}.
\bibitem[{Kaya and Bilge(2019)}]{kaya2019deep}
\bibinfo{author}{M.~Kaya}, \bibinfo{author}{H.~{\c{S}}. Bilge},
\newblock \bibinfo{title}{Deep metric learning: A survey},
\newblock \bibinfo{journal}{Symmetry} \bibinfo{volume}{11}
  (\bibinfo{year}{2019}) \bibinfo{pages}{1066}.
\bibitem[{Wen et~al.(2016)Wen, Zhang, Li, and Qiao}]{wen2016discriminative}
\bibinfo{author}{Y.~Wen}, \bibinfo{author}{K.~Zhang}, \bibinfo{author}{Z.~Li},
  \bibinfo{author}{Y.~Qiao},
\newblock \bibinfo{title}{A discriminative feature learning approach for deep
  face recognition},
\newblock in: \bibinfo{booktitle}{European Conference on Computer Vision},
  \bibinfo{organization}{Springer}, \bibinfo{year}{2016}, pp.
  \bibinfo{pages}{499--515}.
\bibitem[{Liu et~al.(2017)Liu, Wen, Yu, Li, Raj, and Song}]{liu2017sphereface}
\bibinfo{author}{W.~Liu}, \bibinfo{author}{Y.~Wen}, \bibinfo{author}{Z.~Yu},
  \bibinfo{author}{M.~Li}, \bibinfo{author}{B.~Raj}, \bibinfo{author}{L.~Song},
\newblock \bibinfo{title}{{SphereFace}: Deep hypersphere embedding for face
  recognition},
\newblock in: \bibinfo{booktitle}{Proceedings of the IEEE Conference on
  Computer Vision and Pattern Recognition}, \bibinfo{year}{2017}, pp.
  \bibinfo{pages}{212--220}.
\bibitem[{Wang et~al.(2018)Wang, Wang, Zhou, Ji, Gong, Zhou, Li, and
  Liu}]{wang2018cosface}
\bibinfo{author}{H.~Wang}, \bibinfo{author}{Y.~Wang},
  \bibinfo{author}{Z.~Zhou}, \bibinfo{author}{X.~Ji},
  \bibinfo{author}{D.~Gong}, \bibinfo{author}{J.~Zhou},
  \bibinfo{author}{Z.~Li}, \bibinfo{author}{W.~Liu},
\newblock \bibinfo{title}{{CosFace}: Large margin cosine loss for deep face
  recognition},
\newblock in: \bibinfo{booktitle}{Proceedings of the IEEE Conference on
  Computer Vision and Pattern Recognition}, \bibinfo{year}{2018}, pp.
  \bibinfo{pages}{5265--5274}.
\bibitem[{Deng et~al.(2019)Deng, Guo, Xue, and Zafeiriou}]{deng2019arcface}
\bibinfo{author}{J.~Deng}, \bibinfo{author}{J.~Guo}, \bibinfo{author}{N.~Xue},
  \bibinfo{author}{S.~Zafeiriou},
\newblock \bibinfo{title}{{ArcFace}: Additive angular margin loss for deep face
  recognition},
\newblock in: \bibinfo{booktitle}{Proceedings of the IEEE/CVF Conference on
  Computer Vision and Pattern Recognition}, \bibinfo{year}{2019}, pp.
  \bibinfo{pages}{4690--4699}.
\bibitem[{Liu et~al.(2019{\natexlab{a}})Liu, Zhu, Lei, and
  Li}]{liu2019adaptiveface}
\bibinfo{author}{H.~Liu}, \bibinfo{author}{X.~Zhu}, \bibinfo{author}{Z.~Lei},
  \bibinfo{author}{S.~Z. Li},
\newblock \bibinfo{title}{{AdaptiveFace}: Adaptive margin and sampling for face
  recognition},
\newblock in: \bibinfo{booktitle}{Proceedings of the IEEE/CVF Conference on
  Computer Vision and Pattern Recognition}, \bibinfo{year}{2019}{\natexlab{a}},
  pp. \bibinfo{pages}{11947--11956}.
\bibitem[{Liu et~al.(2019{\natexlab{b}})Liu, Deng, Zhong, Wang, Hu, Tao, and
  Huang}]{liu2019fair}
\bibinfo{author}{B.~Liu}, \bibinfo{author}{W.~Deng},
  \bibinfo{author}{Y.~Zhong}, \bibinfo{author}{M.~Wang},
  \bibinfo{author}{J.~Hu}, \bibinfo{author}{X.~Tao},
  \bibinfo{author}{Y.~Huang},
\newblock \bibinfo{title}{Fair loss: Margin-aware reinforcement learning for
  deep face recognition},
\newblock in: \bibinfo{booktitle}{Proceedings of the IEEE/CVF International
  Conference on Computer Vision}, \bibinfo{year}{2019}{\natexlab{b}}, pp.
  \bibinfo{pages}{10052--10061}.
\bibitem[{Huang et~al.(2020)Huang, Wang, Tai, Liu, Shen, Li, Li, and
  Huang}]{huang2020curricularface}
\bibinfo{author}{Y.~Huang}, \bibinfo{author}{Y.~Wang},
  \bibinfo{author}{Y.~Tai}, \bibinfo{author}{X.~Liu},
  \bibinfo{author}{P.~Shen}, \bibinfo{author}{S.~Li}, \bibinfo{author}{J.~Li},
  \bibinfo{author}{F.~Huang},
\newblock \bibinfo{title}{{CurricularFace}: Adaptive curriculum learning loss
  for deep face recognition},
\newblock in: \bibinfo{booktitle}{Proceedings of the IEEE/CVF Conference on
  Computer Vision and Pattern Recognition}, \bibinfo{year}{2020}, pp.
  \bibinfo{pages}{5901--5910}.
\bibitem[{Wang and Deng(2018)}]{wang2018deep}
\bibinfo{author}{M.~Wang}, \bibinfo{author}{W.~Deng},
\newblock \bibinfo{title}{Deep visual domain adaptation: A survey},
\newblock \bibinfo{journal}{Neurocomputing} \bibinfo{volume}{312}
  (\bibinfo{year}{2018}) \bibinfo{pages}{135--153}.
\bibitem[{Long et~al.(2015)Long, Cao, Wang, and Jordan}]{long2015learning}
\bibinfo{author}{M.~Long}, \bibinfo{author}{Y.~Cao}, \bibinfo{author}{J.~Wang},
  \bibinfo{author}{M.~I. Jordan},
\newblock \bibinfo{title}{Learning transferable features with deep adaptation
  networks},
\newblock in: \bibinfo{booktitle}{International Conference on Machine
  Learning}, \bibinfo{organization}{PMLR}, \bibinfo{year}{2015}, pp.
  \bibinfo{pages}{97--105}.
\bibitem[{Sun and Saenko(2016)}]{sun2016deep}
\bibinfo{author}{B.~Sun}, \bibinfo{author}{K.~Saenko},
\newblock \bibinfo{title}{{Deep CORAL}: Correlation alignment for deep domain
  adaptation},
\newblock in: \bibinfo{booktitle}{European Conference on Computer Vision},
  \bibinfo{organization}{Springer}, \bibinfo{year}{2016}, pp.
  \bibinfo{pages}{443--450}.
\bibitem[{Long et~al.(2018)Long, Cao, Cao, Wang, and
  Jordan}]{long2018transferable}
\bibinfo{author}{M.~Long}, \bibinfo{author}{Y.~Cao}, \bibinfo{author}{Z.~Cao},
  \bibinfo{author}{J.~Wang}, \bibinfo{author}{M.~I. Jordan},
\newblock \bibinfo{title}{Transferable representation learning with deep
  adaptation networks},
\newblock \bibinfo{journal}{IEEE Transactions on Pattern Analysis and Machine
  Intelligence} \bibinfo{volume}{41} (\bibinfo{year}{2018})
  \bibinfo{pages}{3071--3085}.
\bibitem[{Ganin and Lempitsky(2015)}]{ganin2015unsupervised}
\bibinfo{author}{Y.~Ganin}, \bibinfo{author}{V.~Lempitsky},
\newblock \bibinfo{title}{Unsupervised domain adaptation by backpropagation},
\newblock in: \bibinfo{booktitle}{International Conference on Machine
  Learning}, \bibinfo{organization}{PMLR}, \bibinfo{year}{2015}, pp.
  \bibinfo{pages}{1180--1189}.
\bibitem[{Tzeng et~al.(2017)Tzeng, Hoffman, Saenko, and
  Darrell}]{tzeng2017adversarial}
\bibinfo{author}{E.~Tzeng}, \bibinfo{author}{J.~Hoffman},
  \bibinfo{author}{K.~Saenko}, \bibinfo{author}{T.~Darrell},
\newblock \bibinfo{title}{Adversarial discriminative domain adaptation},
\newblock in: \bibinfo{booktitle}{Proceedings of the IEEE Conference on
  Computer Vision and Pattern Recognition}, \bibinfo{year}{2017}, pp.
  \bibinfo{pages}{7167--7176}.
\bibitem[{Ma et~al.(2019)Ma, Zhang, and Xu}]{ma2019deep}
\bibinfo{author}{X.~Ma}, \bibinfo{author}{T.~Zhang}, \bibinfo{author}{C.~Xu},
\newblock \bibinfo{title}{Deep multi-modality adversarial networks for
  unsupervised domain adaptation},
\newblock \bibinfo{journal}{IEEE Transactions on Multimedia}
  \bibinfo{volume}{21} (\bibinfo{year}{2019}) \bibinfo{pages}{2419--2431}.
\bibitem[{Yan et~al.(2019)Yan, Li, Wang, Li, Xu, and Zuo}]{yan2019weighted}
\bibinfo{author}{H.~Yan}, \bibinfo{author}{Z.~Li}, \bibinfo{author}{Q.~Wang},
  \bibinfo{author}{P.~Li}, \bibinfo{author}{Y.~Xu}, \bibinfo{author}{W.~Zuo},
\newblock \bibinfo{title}{Weighted and class-specific maximum mean discrepancy
  for unsupervised domain adaptation},
\newblock \bibinfo{journal}{IEEE Transactions on Multimedia}
  \bibinfo{volume}{22} (\bibinfo{year}{2019}) \bibinfo{pages}{2420--2433}.
\bibitem[{Li et~al.(2016)Li, Wang, Shi, Liu, and Hou}]{li2016revisiting}
\bibinfo{author}{Y.~Li}, \bibinfo{author}{N.~Wang}, \bibinfo{author}{J.~Shi},
  \bibinfo{author}{J.~Liu}, \bibinfo{author}{X.~Hou},
\newblock \bibinfo{title}{Revisiting batch normalization for practical domain
  adaptation},
\newblock \bibinfo{journal}{arXiv preprint arXiv:1603.04779}
  (\bibinfo{year}{2016}).
\bibitem[{Qing et~al.(2018)Qing, Zhao, Shi, Chen, Lin, and
  Peng}]{qing2018improve}
\bibinfo{author}{Y.~Qing}, \bibinfo{author}{Y.~Zhao}, \bibinfo{author}{Y.~Shi},
  \bibinfo{author}{D.~Chen}, \bibinfo{author}{Y.~Lin},
  \bibinfo{author}{Y.~Peng},
\newblock \bibinfo{title}{Improve cross-domain face recognition with
  {IBN}-block},
\newblock in: \bibinfo{booktitle}{IEEE International Conference on Big Data
  (BIG DATA)}, \bibinfo{organization}{IEEE}, \bibinfo{year}{2018}, pp.
  \bibinfo{pages}{4613--4618}.
\bibitem[{Nam and Kim(2018)}]{nam2018batch}
\bibinfo{author}{H.~Nam}, \bibinfo{author}{H.-E. Kim},
\newblock \bibinfo{title}{Batch-instance normalization for adaptively
  style-invariant neural networks},
\newblock \bibinfo{journal}{arXiv preprint arXiv:1805.07925}
  (\bibinfo{year}{2018}).
\bibitem[{Qian et~al.(2019)Qian, Jin, Li, Lang, Feng, and Wang}]{qian2019deep}
\bibinfo{author}{C.~Qian}, \bibinfo{author}{Y.~Jin}, \bibinfo{author}{Y.~Li},
  \bibinfo{author}{C.~Lang}, \bibinfo{author}{S.~Feng},
  \bibinfo{author}{T.~Wang},
\newblock \bibinfo{title}{Deep domain adaptation for {Asian} face recognition
  via {Ada-IBN}},
\newblock in: \bibinfo{booktitle}{IEEE International Conference on Multimedia
  \& Expo Workshops (ICMEW)}, \bibinfo{organization}{IEEE},
  \bibinfo{year}{2019}, pp. \bibinfo{pages}{525--530}.
\bibitem[{Jin et~al.(2021)Jin, Lan, Zeng, and Chen}]{jin2021style}
\bibinfo{author}{X.~Jin}, \bibinfo{author}{C.~Lan}, \bibinfo{author}{W.~Zeng},
  \bibinfo{author}{Z.~Chen},
\newblock \bibinfo{title}{Style normalization and restitution for domain
  generalization and adaptation},
\newblock \bibinfo{journal}{arXiv preprint arXiv:2101.00588}
  (\bibinfo{year}{2021}).
\bibitem[{Iandola et~al.(2016)Iandola, Han, Moskewicz, Ashraf, Dally, and
  Keutzer}]{iandola2016squeezenet}
\bibinfo{author}{F.~N. Iandola}, \bibinfo{author}{S.~Han},
  \bibinfo{author}{M.~W. Moskewicz}, \bibinfo{author}{K.~Ashraf},
  \bibinfo{author}{W.~J. Dally}, \bibinfo{author}{K.~Keutzer},
\newblock \bibinfo{title}{{SqueezeNet}: {AlexNet}-level accuracy with 50x fewer
  parameters and $<$0.5{MB} model size},
\newblock \bibinfo{journal}{arXiv preprint arXiv:1602.07360}
  (\bibinfo{year}{2016}).
\bibitem[{Genevay et~al.(2018)Genevay, Peyr{\'e}, and
  Cuturi}]{genevay2018learning}
\bibinfo{author}{A.~Genevay}, \bibinfo{author}{G.~Peyr{\'e}},
  \bibinfo{author}{M.~Cuturi},
\newblock \bibinfo{title}{Learning generative models with {S}inkhorn
  divergences},
\newblock in: \bibinfo{booktitle}{International Conference on Artificial
  Intelligence and Statistics}, \bibinfo{organization}{PMLR},
  \bibinfo{year}{2018}, pp. \bibinfo{pages}{1608--1617}.
\bibitem[{Learned-Miller(2014)}]{LFWTechUpdate}
\bibinfo{author}{G.~B. H.~E. Learned-Miller}, \bibinfo{title}{Labeled {F}aces
  in the {W}ild: Updates and New Reporting Procedures},
  \bibinfo{type}{Technical Report} \bibinfo{number}{UM-CS-2014-003}, University
  of Massachusetts, Amherst, \bibinfo{year}{2014}.
\bibitem[{Wolf et~al.(2011)Wolf, Hassner, and Maoz}]{wolf2011face}
\bibinfo{author}{L.~Wolf}, \bibinfo{author}{T.~Hassner},
  \bibinfo{author}{I.~Maoz},
\newblock \bibinfo{title}{Face recognition in unconstrained videos with matched
  background similarity},
\newblock in: \bibinfo{booktitle}{CVPR 2011}, \bibinfo{organization}{IEEE},
  \bibinfo{year}{2011}, pp. \bibinfo{pages}{529--534}.
\bibitem[{Cao et~al.(2018)Cao, Shen, Xie, Parkhi, and
  Zisserman}]{cao2018vggface2}
\bibinfo{author}{Q.~Cao}, \bibinfo{author}{L.~Shen}, \bibinfo{author}{W.~Xie},
  \bibinfo{author}{O.~M. Parkhi}, \bibinfo{author}{A.~Zisserman},
\newblock \bibinfo{title}{{VGGFace2}: A dataset for recognising faces across
  pose and age},
\newblock in: \bibinfo{booktitle}{13th IEEE international conference on
  automatic face \& gesture recognition (FG 2018)},
  \bibinfo{organization}{IEEE}, \bibinfo{year}{2018}, pp.
  \bibinfo{pages}{67--74}.
\bibitem[{Yi et~al.(2014)Yi, Lei, Liao, and Li}]{yi2014learning}
\bibinfo{author}{D.~Yi}, \bibinfo{author}{Z.~Lei}, \bibinfo{author}{S.~Liao},
  \bibinfo{author}{S.~Z. Li},
\newblock \bibinfo{title}{Learning face representation from scratch},
\newblock \bibinfo{journal}{arXiv preprint arXiv:1411.7923}
  (\bibinfo{year}{2014}).
\bibitem[{Sengupta et~al.(2016)Sengupta, Chen, Castillo, Patel, Chellappa, and
  Jacobs}]{sengupta2016frontal}
\bibinfo{author}{S.~Sengupta}, \bibinfo{author}{J.-C. Chen},
  \bibinfo{author}{C.~Castillo}, \bibinfo{author}{V.~M. Patel},
  \bibinfo{author}{R.~Chellappa}, \bibinfo{author}{D.~W. Jacobs},
\newblock \bibinfo{title}{Frontal to profile face verification in the wild},
\newblock in: \bibinfo{booktitle}{2016 IEEE Winter Conference on Applications
  of Computer Vision (WACV)}, \bibinfo{organization}{IEEE},
  \bibinfo{year}{2016}, pp. \bibinfo{pages}{1--9}.
\bibitem[{Moschoglou et~al.(2017)Moschoglou, Papaioannou, Sagonas, Deng,
  Kotsia, and Zafeiriou}]{moschoglou2017agedb}
\bibinfo{author}{S.~Moschoglou}, \bibinfo{author}{A.~Papaioannou},
  \bibinfo{author}{C.~Sagonas}, \bibinfo{author}{J.~Deng},
  \bibinfo{author}{I.~Kotsia}, \bibinfo{author}{S.~Zafeiriou},
\newblock \bibinfo{title}{{AgeDB}: the first manually collected, in-the-wild
  age database},
\newblock in: \bibinfo{booktitle}{Proceedings of the IEEE Conference on
  Computer Vision and Pattern Recognition Workshops}, \bibinfo{year}{2017}, pp.
  \bibinfo{pages}{51--59}.
\bibitem[{Hsu et~al.(2018)Hsu, Wu, Wan, Wong, and Lee}]{hsu2018quatnet}
\bibinfo{author}{H.-W. Hsu}, \bibinfo{author}{T.-Y. Wu},
  \bibinfo{author}{S.~Wan}, \bibinfo{author}{W.~H. Wong},
  \bibinfo{author}{C.-Y. Lee},
\newblock \bibinfo{title}{Quatnet: Quaternion-based head pose estimation with
  multiregression loss},
\newblock \bibinfo{journal}{IEEE Transactions on Multimedia}
  \bibinfo{volume}{21} (\bibinfo{year}{2018}) \bibinfo{pages}{1035--1046}.
\bibitem[{Zhang et~al.(2016)Zhang, Zhang, Li, and Qiao}]{zhang2016joint}
\bibinfo{author}{K.~Zhang}, \bibinfo{author}{Z.~Zhang},
  \bibinfo{author}{Z.~Li}, \bibinfo{author}{Y.~Qiao},
\newblock \bibinfo{title}{Joint face detection and alignment using multitask
  cascaded convolutional networks},
\newblock \bibinfo{journal}{IEEE Signal Processing Letters}
  \bibinfo{volume}{23} (\bibinfo{year}{2016}) \bibinfo{pages}{1499--1503}.
\bibitem[{Chen et~al.(2018)Chen, Liu, Gao, and Han}]{chen2018mobilefacenets}
\bibinfo{author}{S.~Chen}, \bibinfo{author}{Y.~Liu}, \bibinfo{author}{X.~Gao},
  \bibinfo{author}{Z.~Han},
\newblock \bibinfo{title}{{MobileFaceNets}: Efficient {CNNs} for accurate
  real-time face verification on mobile devices},
\newblock in: \bibinfo{booktitle}{Chinese Conference on Biometric Recognition},
  \bibinfo{organization}{Springer}, \bibinfo{year}{2018}, pp.
  \bibinfo{pages}{428--438}.
\bibitem[{Gretton et~al.(2012)Gretton, Sejdinovic, Strathmann, Balakrishnan,
  Pontil, Fukumizu, and Sriperumbudur}]{gretton2012optimal}
\bibinfo{author}{A.~Gretton}, \bibinfo{author}{D.~Sejdinovic},
  \bibinfo{author}{H.~Strathmann}, \bibinfo{author}{S.~Balakrishnan},
  \bibinfo{author}{M.~Pontil}, \bibinfo{author}{K.~Fukumizu},
  \bibinfo{author}{B.~K. Sriperumbudur},
\newblock \bibinfo{title}{Optimal kernel choice for large-scale two-sample
  tests},
\newblock in: \bibinfo{booktitle}{Advances in neural information processing
  systems}, \bibinfo{organization}{Citeseer}, \bibinfo{year}{2012}, pp.
  \bibinfo{pages}{1205--1213}.
\bibitem[{He et~al.(2016)He, Zhang, Ren, and Sun}]{he2016deep}
\bibinfo{author}{K.~He}, \bibinfo{author}{X.~Zhang}, \bibinfo{author}{S.~Ren},
  \bibinfo{author}{J.~Sun},
\newblock \bibinfo{title}{Deep residual learning for image recognition},
\newblock in: \bibinfo{booktitle}{Proceedings of the IEEE Conference on
  Computer Vision and Pattern Recognition}, \bibinfo{year}{2016}, pp.
  \bibinfo{pages}{770--778}.

\end{thebibliography}





\end{document}